%% file: acl_latex.tex
\pdfoutput=1
\documentclass[11pt]{article}

\usepackage[table,xcdraw]{xcolor}
\definecolor{blueoriginal}{HTML}{00046E}
\definecolor{blueimplicature}{HTML}{147BF5}
\definecolor{rednegation}{HTML}{F24822}

\usepackage[]{acl}
\usepackage{fontawesome5}

\usepackage{times}
\usepackage{latexsym}
\usepackage{amsfonts}
\usepackage{amsmath}
\usepackage{graphicx,subfigure,subcaption}
\usepackage{booktabs}
\usepackage{multirow}
\usepackage{url}
\usepackage{tabularx}
\usepackage{makecell}
\usepackage{tabularray}
\usepackage[vlined,ruled,commentsnumbered,linesnumbered]{algorithm2e}

\usepackage[colorinlistoftodos,prependcaption,textsize=tiny]{todonotes}
\PassOptionsToPackage{table}{xcolor}

\DeclareMathOperator*{\argmin}{argmin}

\newcolumntype{s}{>{\hsize=0.62\hsize}X}
\newcolumntype{t}{>{\hsize=.60\hsize}X}

\usepackage[T1]{fontenc}
\usepackage[utf8]{inputenc}

\usepackage{microtype}

\usepackage{inconsolata}

\title{
Can Your Model Tell a Negation from an Implicature?\\Unravelling Challenges With Intent Encoders
}

\author{
    Yuwei Zhang$^{\dag}$\thanks{~~Work conducted as in an intern/employee of Amazon.} \quad
    Siffi Singh$^\ddag$ \quad
    Sailik Sengupta$^\ddag$ \quad
    Igor Shalyminov$^\ddag$ \\[0.3em]
    {\bf Hang Su$^\ddag$ \quad Hwanjun Song$^{\P*}$ \quad
    Saab Mansour$^\ddag$}\\[0.8em]
    $^\dag$University of California, San Diego \quad
    $^\ddag${\color{orange} \faAmazon}WS AI Labs \quad
    $^\P$KAIST, Republic of Korea\\[0.3em]
    \faIcon[regular]{envelope} {\tt \{siffis, sailiks\}@amazon.com}
}

\begin{document}
\maketitle
\begin{abstract}
\input{sections/0_abstract}
\end{abstract}

\input{sections/1_intro}

\input{sections/2_rel}

\input{sections/3_prelim}

\input{sections/4_eval}

\input{sections/5_train}

\input{sections/6_exp}

\input{sections/7_con}

\input{sections/lim}

\input{sections/ack}

\bibliography{custom}

\appendix

\input{sections/app}

\end{document}

%% file: sections/0_abstract.tex
Conversational systems often rely on embedding models for intent classification and intent clustering tasks. The advent of Large Language Models (LLMs), which enable {\em instructional embeddings} allowing one to adjust semantics over the embedding space using prompts, are being viewed as a panacea for these downstream conversational tasks. However, traditional evaluation benchmarks rely solely on task metrics that don't particularly measure gaps related to semantic understanding. Thus, we propose an intent semantic toolkit that gives a more holistic view of intent embedding models by considering three tasks-- (1) intent classification, (2) intent clustering, and (3) a novel triplet task. The triplet task gauges the model's understanding of two semantic concepts paramount in real-world conversational systems-- {\em negation} and {\em implicature}. We observe that current embedding models fare poorly in semantic understanding of these concepts. To address this, we propose a pre-training approach to improve the embedding model by leveraging augmentation with data generated by an auto-regressive model and a contrastive loss term. Our approach improves the semantic understanding of the intent embedding model on the aforementioned linguistic dimensions while slightly effecting their performance on downstream task metrics.

%% file: sections/1_intro.tex
\section{Introduction}\label{sec:intro}

Conversational systems use intent embedding models to encode input utterances into vectors that are used to understand intent semantics for few-shot intent classification and/or intent discovery~\cite{zhang-etal-2021-effectiveness-pre,ma-etal-2022-effectiveness,sung2023pretraining}. The intent classification task leverages a pre-defined distance metric to find the nearest intent class/instances to the test utterance in the embedding space~\cite{vinyals2016matching,snell2017prototypical,dopierre-etal-2021-protaugment}, while intent discovery considers a clustering algorithm on top of multiple utterance embeddings to detects novel intent clusters. While these applications have been studied separately, a common underlying assumption is the embedding space encodes semantics (and a distance metric) between utterances and intents. To achieve this, various approaches have been proposed such as supervised pre-training on labeled utterance data belonging to a wide range of intents~\cite{zhang-etal-2021-effectiveness-pre,zhang-etal-2022-fine} or by combining pseudo intent names~\cite{sung2023pretraining,mueller-etal-2022-label}.
%A distance metric can then be applied to measure similarity between utterances or intents.
%Such a model is often used in a conversational agent to perform intent-related tasks due to its low computational costs compared with Large Language Models.
% Furthermore, these models are usually designed for a general purpose intent-related applications that can work on multiple domains.
%The only computation required is to encode the utterances with a transformer-based model and perform inner product or calculate euclidean distances between vectors.

However, the evaluation of these embedding models only consider existing conversational benchmarks (and task metrics) that lack dedicated test data necessary to evaluate gaps in semantic understanding. In this paper, we consider a complementary evaluation approach that tries to understand how well these embedding models capture the semantics of two common linguistic phenomenon seen in real-world conversational systems: \textbf{negation} -- negation semantics alongside an explicitly mentioned intent utterance (e.g. {\em No, I don't want you to play music!}) and \textbf{implicature} -- utterances indirectly hinting at an intent that might require some reasoning steps (e.g. {\em I feel like dancing} $\implies$ {\em play some music}). For this purpose, we propose an {\em Intent Semantics Toolkit} that includes challenging test splits for existing classification \& clustering tasks and a novel triplet task. The triplet task consider an utterance triplet $\langle$original, implicature, negation$\rangle$ and evaluates if the implicature utterance is closer to the original utterance in the embedding space as compared to negation. The implicature and negation utterances for the test data are generated using two novel prompt designs for ChatGPT followed by human-in-the-loop quality control mechanisms.
%Our work differs from many existing works that consider evaluation on various linguistic phenomenon in two ways-- (1) we don't try to couple semantic understanding  they did not consider it in the context of (intent) embedding models~\cite{marelli-etal-2014-sick,jeretic-etal-2020-natural,sengupta-etal-2021-robustness,ruis2022large,cooper-stickland-etal-2023-robustification}.

\begin{figure*}[t]
    \centering
    \includegraphics[width=0.88\textwidth]{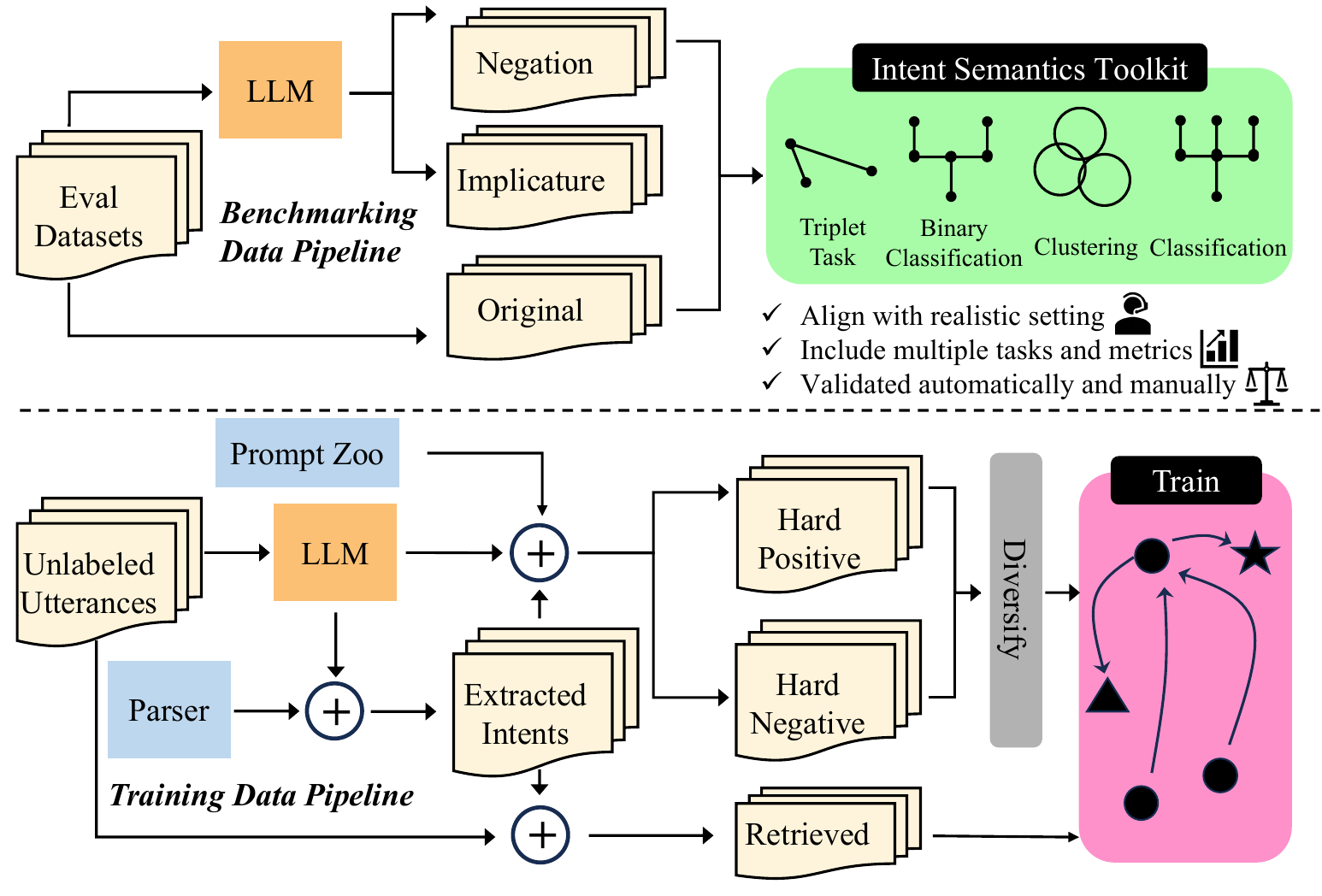}
    \caption{Intent Semantics Toolkit (top) to benchmark embedding model semantic concepts capabilities and training pipeline (bottom) to synthesize training data for improved semantic concepts understanding. For benchmarking data, we prompt the LLM to generate negation and implicature from original utterances, which are validated by both automatic and manual quality control. For training data, we first extract intents from unlabeled utterances, then, we generate hard examples using the LLM, which will be combined with retrieved utterances for fine-tuning.}
    %\caption{Overall framework for generating our proposed Intent Semantics Toolkit (top) and hard examples during model fine-tuning (bottom). For benchmarking data, we prompt the LLM to generate negation and implicature from original utterances, which are validated by both automatic and manual quality control. Combined with original utterances, we create the evaluation toolkit with $4$ different tasks. For training data, we first extract intents from unlabeled utterances, then, we generate hard examples using the LLM, which will be combined with retrieved utterances for fine-tuning.}
    \label{fig:framework}
    \vspace{-4.5mm}
\end{figure*}
\begin{figure}[t]
    \centering
    %l, b, r, t
    \includegraphics[width=0.49\textwidth,trim={4cm 4.2cm 4cm 4.5cm},clip]{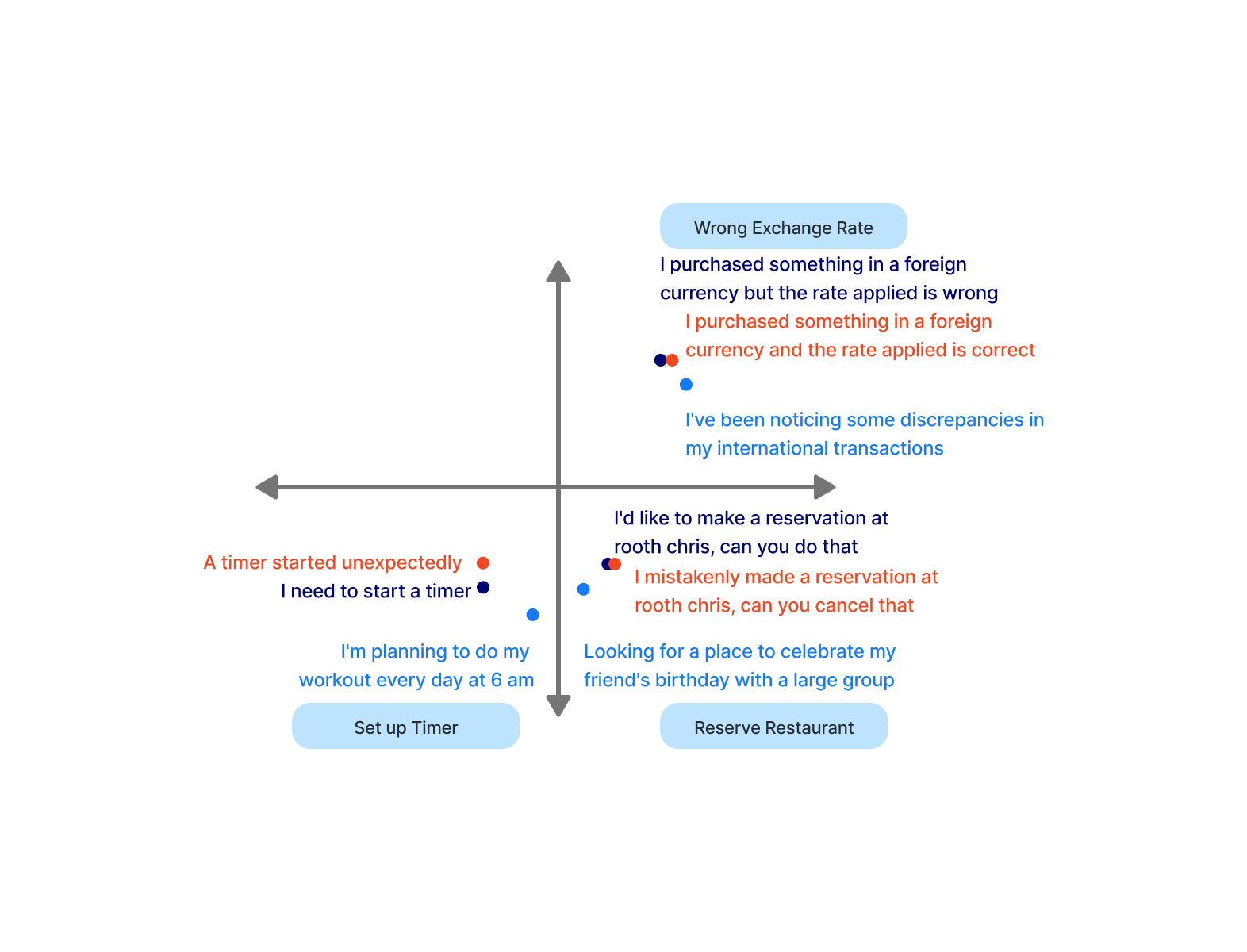}
    % Failure cases of \texttt{instructor-large} on the triplet task with cosine distance and the prompt \textit{Represent the purpose for retrieval}. The {\color{fblue} implicature} utterances are further to the original utterances than {\color{forange} negation} ones.
    \caption{
    The {\tt instructor-large} embeds {\color{blueoriginal} original} utterances further away from semantically similar {\color{blueimplicature} implicature} utterances and closer to semantically dissimilar {\color{rednegation} negation} utterances, a failure mode for the triplet task (as seen in the tSNE projection space).
    }
    \label{fig:challenges}
    \vspace{-6mm}
\end{figure}

The recent popularity of embeddings derived from Large Language Models and the possibility of prompt-based encoding give an impression of semantic understanding, making them seem like an ideal candidate for the aforementioned intent identification tasks. Our proposed {\em Intent Semantics Toolkit} indicates that current representation of the negation and implicature utterances are far from perfect (see \autoref{fig:challenges}).

To improve an embedding model's semantic understanding on the aforementioned linguistic phenomena, we consider a fine-tuning approach that leverages LLM-generated positive (related to an intent) and negative (unrelated to an intent) utterances for augmentation alongside a contrastive loss objective. Our ablations highlight the need for both kinds of utterances for augmentation and our best model consistently outperforms the original embedding models on the triplet task. Observing different magnitudes of improvement on the different tasks, we dive deep to understand correlation between the various tasks and highlight some negative interactions. This highlights that approaches to improve embedding models on semantic understanding and downstream task might need to consider trade-offs; this is in line with several previous works~\cite{marelli-etal-2014-sick,jeretic-etal-2020-natural,sengupta-etal-2021-robustness,cooper-stickland-etal-2023-robustification}. To summarize, our contributions are three-fold:
(1) We identify two linguistic challenges that are commonly observed in real-world intent detection systems, but are overlooked in the literature: negation and implicature.\footnote{We note that these phenomenon also have equivalence in the context of reasoning.}
(2) We devise Intent Semantics Toolkit that includes a novel triplet task and exposes the shortcomings of intent embedding models on semantic understanding of negation and implicature. We propose prompting strategies to generate evaluation data with ChatGPT and consider human-in-the-loop quality control.
(3) We explore fine-tuning approaches with automatically generated utterances for data augmentation and interpret the semantic dimensions of implicature and negation as positive and negative examples in the contrastive learning loss. The results show that generated utterances can help improve the performance on the triplet task.

%% file: sections/2_rel.tex
\section{Related Work}\label{sec:rel}
% listing important related works for now
% for papers published in ACL conferences, they are already integrated in anthology.bib, search for Bibkey
% for other papers please add in custom.bib
% \begin{table*}
% \centering
% \begin{tabular}{lll}
% \hline
% \textbf{Output} & \textbf{natbib command} & \textbf{Old ACL-style command}\\
% \hline
% \citep{su-etal-2023-one} & \verb|\citep| & \verb|\cite| \\
% \citealp{su-etal-2023-one} & \verb|\citealp| & no equivalent \\
% \citet{su-etal-2023-one} & \verb|\citet| & \verb|\newcite| \\
% \citeyearpar{su-etal-2023-one} & \verb|\citeyearpar| & \verb|\shortcite| \\
% \hline
% \end{tabular}
% \caption{\label{citation-guide}
% Citation commands supported by the style file.
% The style is based on the natbib package and supports all natbib citation commands.
% It also supports commands defined in previous ACL style files for compatibility.
% }
% \end{table*}

\noindent \textbf{Sentence embedding evaluation.}
Sentence embeddings are used in various downstream applications, especially where efficiency is important \cite{reimers-gurevych-2019-sentence,su-etal-2023-one}. For instance, semantic search of relevant document from a large vector database based on a similarity metric~\cite{nguyen2016msmarco,thakur2021beir}, clustering a set of `unorganized' documents based on the semantics ~\cite{macqueen1967some,ester1996density}, and/or classification of an input at test time to the semantically closest class seen during training \cite{vinyals2016matching,snell2017prototypical,conneau-kiela-2018-senteval}.
Recent works have also proposed a benchmark that unifies the evaluation of sentence embeddings across the various tasks~\cite{muennighoff-etal-2023-mteb}.
Along these lines, \citet{liu2023sentbench} proposes a triplet evaluation task that examines similarity of triplets sampled from existing class labels. In our work, we also propose a triplet evaluation task, but consider examining triplets based on dimensions related to language semantics such as implication and negation.

\noindent \textbf{Conversational implicature.}
% Conversational implicature is often conveyed with an utterance to imply something without literally mentioning it~\cite{grice1975logic}. It is the non-conventional meaning of the utterance and is the assumption made in a certain context~\cite{recanati1989pragmatics}, which can be used to efficiently communicate, maintain social relations or mislead without lying~\cite{zalta1995stanford}. For instance, the utterance ``I have not had breakfast today'' can imply that the person is hungry even though it is not literally being said.
% In the domain of natural language processing, it is often implicitly mentioned in QA settings when a question needs to be clarified~\cite{aliannejadi-etal-2021-building,sun-etal-2022-conditionalqa,zhou-etal-2022-think}. There are also works claiming that current NLI language models or even LLM do not have enough understanding of implicature~\cite{jeretic-etal-2020-natural,ruis2022large}. In this work, we brought implicature to intent detection scenarios and we are especially interested in the capabilities of embedding models.
In conversation, an agent can often imply an intent via an utterance that doesn't explicitly specify the intent, but rather hints at it~\cite{grice1975logic,recanati1989pragmatics,zalta1995stanford}. For instance, the utterance {\em  I have not had breakfast today} can imply that the person is hungry and could hint at a particular intent (eg. {\em order food}) depending on the context (eg. saying this to a hotel operator).
Previous works have shown that current language models do not have enough understanding of implicature~\cite{jeretic-etal-2020-natural,ruis2022large}. In this work, we consider implicature for intent detection scenarios and are especially interested in the capabilities of embedding models.

%% file: sections/3_prelim.tex
\section{Preliminary}\label{sec:prelim}

% --- Rewrote this paragraph below by breaking it down into semantically meaningful chunks ---
% An intent embedding model encodes input utterances onto a vector space where semantic similarities to an intent can be computed via a certain distance metric. For instance, consider an utterance $u_i$ is encoded as the embedding $f(u_i)$, where parameters of $f$ are pre-trained. Now, consider a distance metric $D$ used to measure the distance between a pair of embedded utterances $u_i$ and $u_j$. In this paper, we will consider $D$ to be the cosine distance $D(f(u_i),f(u_j)) = 1 - \frac{f(u_i)^T f(u_j)}{||f(u_i)|| ||f(u_j)||}$ as it is commonly used for clustering and classification tasks. In practice, intent embeddings are deployed to discover new intents from unlabeled customer utterances~\citep{zhang-etal-2022-new,gung2023intent} or perform classification in few-/zero- shot settings~\cite{snell2017prototypical,dopierre-etal-2021-protaugment,sung2023pretraining}.

% Define embedding models
A model ($f$) for embedding encodes input sentences ($u_i$) onto a continuous vector space ($f(u_i)$). For a good embedding model, a distance metric $D$ over embeddings (say $f(u_i)$ and $f(u_j)$) should be able to capture some notion of semantic similarity that in turn empowers downstream applications such as clustering and classification. In this paper, we will consider $D$ to be the widely popular cosine distance, i.e. $D(f(u_i),f(u_j)) = 1 - \frac{f(u_i)^T f(u_j)}{||f(u_i)|| ||f(u_j)||}$.

% Introduce the use of embedding models for problems with intents
While these embedding models are common in several applications, we will concentrate on intent embedding models, where input sentences are user utterances (e.g. {\em I would like to order food}) and intents (e.g. {\em order food, cancel order,} etc.) comprise of semantic clusters represented by a set of similar user utterances. In such contexts, embedding models have been used to perform few/zero-shot classification of test-time utterances to a set-of predefined intents~\cite{snell2017prototypical,dopierre-etal-2021-protaugment,sung2023pretraining} or to discover novel intent classes given a set of unlabelled user utterances~\citep{zhang-etal-2022-new,gung2023intent}.

% Talk about chellenges related to language semantices for scenarios with intents
Despite its success, we observe that these intent embedding models often fail to capture nuanced language semantics that are common in real-world conversational agents. While previous work has shown such failures are common in the context of real-world noise \cite{sengupta-etal-2021-robustness,cooper-stickland-etal-2023-robustification}, we focus on two semantic aspects-- (1) \textbf{Negation}: utterances that explicitly express no interest towards a particular intent (e.g. {\em I don't want to order a drink.}),
and (2) \textbf{Implicature}: utterances that do not explicitly convey an intent but imply it (e.g. {\em I am hungry} $\rightarrow$ {\em order\_food}). \autoref{fig:challenges} shows that intent embedding models embed negations (that clearly disregards an intent) closer to intent utterances than implicatures (that implies an intent). In the real-world, we observe negation often occurs when a system incorrectly directs a customer towards an intent (and the customer has to explicitly mention they didn't intend it), while implicature is common due to a customer's incomplete knowledge about the potential functionality supported by a chatbot.

%% file: sections/4_eval.tex
\section{Intent Semantics Toolkit}\label{sec:eval}

In this section, we first introduce the four evaluation tasks in our toolkit that analyze the semantic understanding capabilities of intent embedding models (\S\ref{sec:eval_task}). Then, we consider the challenge of obtaining data related to negation and implicature (\S\ref{sec:prelim}) and quality assurance procedures to ensure a high-quality test set (\S\ref{sec:eval_qual}). Eventually, we show evaluation of SOTA models on our dataset (\S\ref{sec:eval_res}).

% In this section, we first present an evaluation toolkit that includes $4$ different tasks to comprehensively analyze the semantic understanding capabilities of intent embedding models (Section~\ref{sec:eval_task}). After that, we generate high-quality data for the two challenges mentioned in Section~\ref{sec:prelim} with LLMs in order to accomplish the evaluation toolkit (Section~\ref{sec:eval_gen}). Finally, we provide quality controls for generated data with both automatic and manual methods (Section~\ref{sec:eval_qual}).

\subsection{Evaluation Tasks}\label{sec:eval_task}

\paragraph{Triplet Task}
A triplet task is composed of three utterances $\{u_i,u_i^p,u_i^n\}$, where $u_i$ ({\em I want to order a pizza}) is an utterance belonging to intent $i$ ({\em order\_food}), $u_i^n$ is the negation of $u_i$ ({\em I don't want to order pizza}), and $u_i^p$ is either another utterance of the same intent ({\em I need pizza}) or an implicature one ({\em I am really hungry}). For each triplet, we expect the negated utterance $u_i^n$ to be embedded further away from $u_i$ than $u_i^p$, i.e. $D(f(u_i),f(u_i^p))<D(f(u_i),f(u_i^n))$. As we show later, this is often difficult for embedding models that may focus on other surface-form aspects of utterances rather than nuanced semantic understanding of intents. We calculate success among a set of $N_T$ triplets as,%
\begin{equation}
\small
    \frac{1}{N_T}\sum_{i=1}^{N_T} \mathbb{I} (D(f(u_i),f(u_i^p))<D(f(u_i),f(u_i^n)))
\end{equation}%
In principle, we can also define another success rate by interchanging $u_i$ and $u_i^p$ as follows,%
\begin{equation}
\small
    \frac{1}{N_T}\sum_{i=1}^{N_T} \mathbb{I} (D(f(u_i^p),f(u_i))<D(f(u_i^p),f(u_i^n)))
\end{equation}%
Given $u_i^p$ and $u_i^n$ are not direct negations of one another, the value of $D(f(u_i^p), f(u_i^n))$ in (2) is expected to be higher than $D(f(u_i),f(u_i^n)))$ in (1) above. Hence, (2) is a more relaxed success criterion.
%The second definition has a more relaxed requirement since it does not require the negation utterance to be directly further away from its own original form.
Thus, we denote the (1) as $T_{\text{hard}}$ and (2) as $T_{\text{easy}}$ hereafter. We will show two cases ``Ori-Ori'' or ``Ori-Imp'' where $u^p_i$ is from either the original test set or the implicature set.

\paragraph{Binary Classification}
In the binary classification task, an utterance $u_i$ needs to be classified into either original intent $i$ or a negated intent class$\neg i$. Thus, success implies $D(f(u_i),f(i))<D(f(u_i),f(\neg i))$. And the success rate is calculated among a set of $N_B$ utterances,
\begin{equation}
\small
    \frac{1}{N_B} \sum_{i=1}^{N_B} \mathbb{I} (D(f(u_i),f(i))<D(f(u_i),f(\neg i)))
\end{equation}
We compute success rates for three different sets of utterances-- original and implicature (closer to $f(i)$) and negation (closer to $f(\neg i)$).

\paragraph{Clustering}
The input to the task is a set of unlabeled utterances $\{u_i\}$. A clustering algorithm $\mathcal{C}$ then takes the embeddings $\{f(u_i)\}$ as inputs and outputs clustering indices $\{q_i\}$ where number of clusters $k$ is first specified. Since the permutation of $\{q_i\}$ might be different from labels, we measure \textit{normalized mutual information} (NMI) \cite{estevez2009normalized} between them as our metric. We show results on both original and implicature sets with two clustering algorithms: $k$-means~\cite{macqueen1967some} and agglomerative clustering~\cite{nielsen2016hierarchical}.

\paragraph{Multi-class Classification}
% \noindent\textbf{Multi-class Classification.}
We adopt ProtoNet~\cite{snell2017prototypical} as the classifier for few-/zero- shot classification aligning with previous works~\cite{dopierre-etal-2021-protaugment,sung2023pretraining}. It takes as inputs a training set $\{\tilde{u}_j,\tilde{y}_j\}$ and label names $\{l_c\}$. A class prototype is first calculated for each class by averaging the embeddings of utterances of that intent class and the label name,
\begin{equation}
    p_c = \frac{1}{N_c + 1} \{[\sum_{\tilde{y}_j = c} f(\tilde{u}_j)] + f(l_c)\}
\end{equation}
During test time, we simply find the prototype that is closest to the test utterance $\argmin_{c} D(p_c,f(u_i))$.
Different from previous works, we discard episodic training and always use full categories within the dataset instead of random $M$ ways. We argue that such a setting is more realistic and show results on both original and implicature sets with $0/10$-shots.

\subsection{Data Generation}\label{sec:eval_gen}
In this section, we show data generation workflows with ChatGPT\footnote{Both \texttt{gpt-4-0613} and \texttt{gpt-3.5-turbo-0613} are used.} models (in order to achieve high quality) on test splits of $3$ intent-classification benchmarks: BANKING77~\cite{casanueva-etal-2020-efficient}, HWU64~\cite{Liu2021} and CLINC150~\cite{larson-etal-2019-evaluation}. Across all datasets, we needed to make the intent names more informative (\textit{e.g.} ``mpg'' to ``check car mpg'') and focused on four dimensions when optimizing prompts for the LLM-generation:
(R1) \textbf{faithful:} The generated utterances correctly conveys/negates the target intent,
(R2) \textbf{realistic:} The generated utterances can be said by a real-world customer,
(R3) \textbf{diverse:} The generated utterances are diverse enough for evaluation on a large scale, and
(R4) \textbf{reproducible:} The temperature was set to $0$ to guarantee reproducibility.

\noindent\textbf{Negation.} Directly negating an utterance without a target intent has lower guarantee on generation faithfulness. For instance, consider the utterance ``\textit{are there restrictions for carry-ons on Delta}'' that belongs to the intent ``\textit{carry on}''. The generated negation utterance is ``\textit{are there no restrictions for carry-ons on delta}'' with \texttt{gpt-4-0613} and temperature set to $0$, which still belongs to the same intent. In order to reduce the task complexity for LLMs, we first manually write a few negated intents for each original intent. And then we instruct the LLM to directly modify original utterances according to a randomly sampled negated intent corresponding to the original one. To further increase the generation quality, we provide $6$ in-context examples in the prompt manually written by humans. We choose \texttt{gpt-4-0613} for negation.

\noindent\textbf{Implicature.} Implicature utterances are completely generated by LLMs without original utterances provided. In order to achieve diversity and realisticness, we first ask an LLM to brain-storm $10$ scenarios that a customer may encounter scenarios for a certain intent, where the scenario may contain various roles and situations. We choose \texttt{gpt-3.5-turbo-0613} for this part since the preliminary results show sufficient quality. Then, we use another LLM to generate $3$ utterances for each scenario. In addition, we provide the definitions of implicature and $3$ manually written in-context examples for higher quality. We choose \texttt{gpt-4-0613} for the latter part.

% \textcolor{red}{Prompt and examples coming soon.}

\subsection{Data Quality Control}\label{sec:eval_qual}
\begin{figure}[t]
    \centering
    \includegraphics[width=0.4\textwidth]{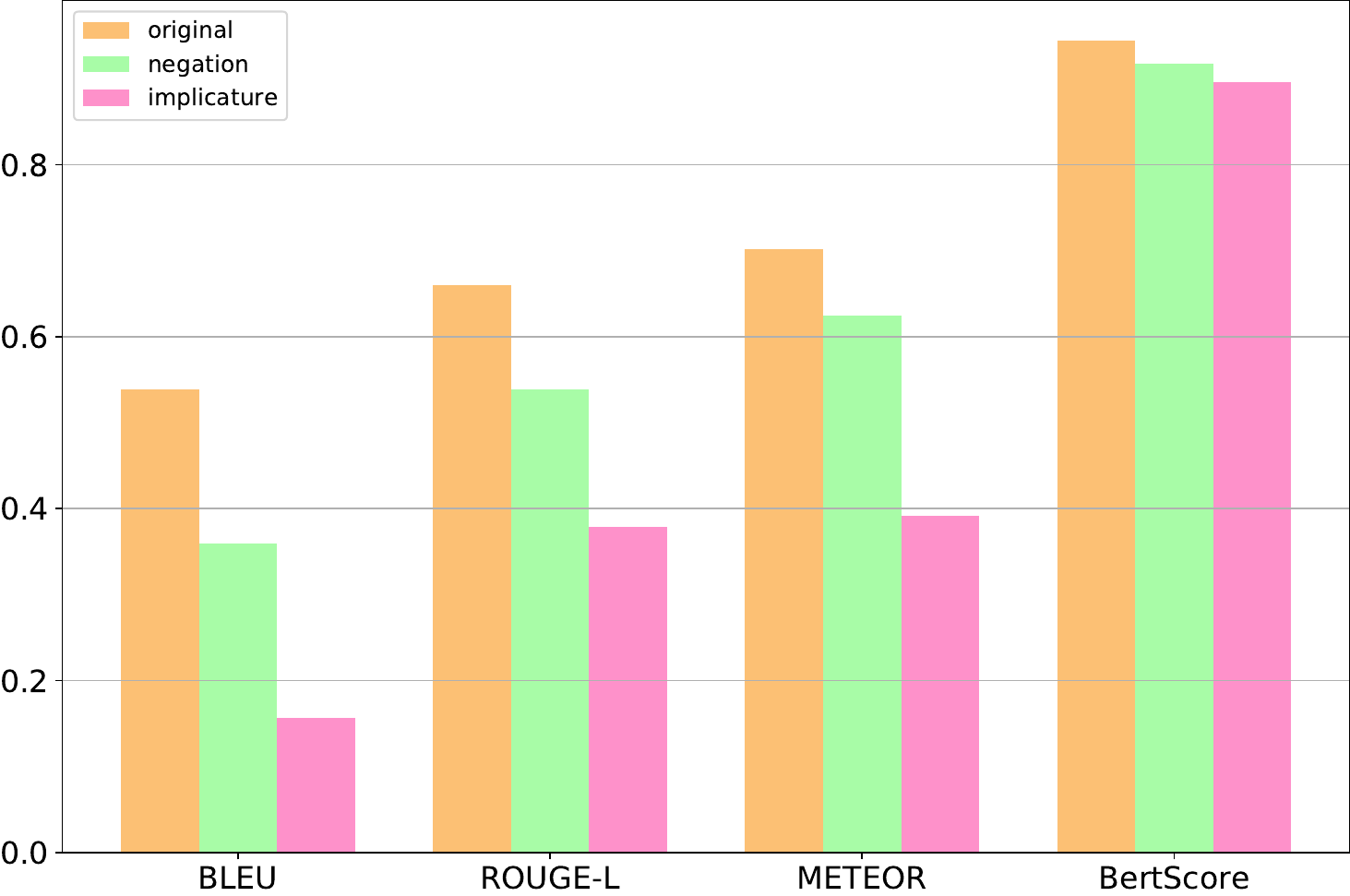}
    \caption{\small Similarity metrics between the training data and the original, and (generated) negation and implicature test splits. Averaged results show the negation utterances are closer to the original ones on surface form than the implicature ones.}
    % \caption{Automatic evaluation. We show several metrics using training data as a reference. The results show that our generated negation utterances are close to the original ones on surface form than the implicature ones.}
    \label{fig:eval_qual_auto}
    % \vspace{-6mm}
\end{figure}

We first employ automatic metrics to understand the generated data. To achieve this, we use training set as reference and calculate BLEU~\cite{papineni-etal-2002-bleu}, ROUGE-L~\cite{lin-2004-rouge}, METEOR~\cite{banerjee-lavie-2005-meteor} and BertScore~\cite{Zhang2020BERTScore}. We show the measures on three sets separately in Figure~\ref{fig:eval_qual_auto}. As expected, original set has the largest vocabulary overlap with training set, probably because the original utterances are generated in the same style with lack of variability. Negation set contains fewer vocabulary overlap but still significantly higher than implicature set, which demonstrates they are similar to original ones in surface form.

However, merely measuring vocabulary overlap does not necessarily guarantee the faithfulness of generated data. Hence, we conduct human evaluation centered around two questions:
(1) ``\textit{Can the utterance imply the intent?}''
(2) ``\textit{If yes, is it conveyed explicitly?}''
We first sample $180$ utterances containing $60$ from original, $60$ from negation and $60$ from implicature. 8 human annotators from our internal team (with fluent English skills) provided binary answers for these two questions in a sequential manner\footnote{We also allow them to annotate ``unsure''.}. We show the annotation guidelines in Figure~\ref{tab:app_annotation}. For each utterance, $3$ annotations are gathered, and those without $3$ annotations are filtered out. We take the majority as the final decision. In Table~\ref{tab:eval_human_breakdown}, we show the results for all three sets. Both negation and implicature generations are indeed faithful. Furthermore, $88.46\%$ of original set are annotated as explicit while $62.50\%$ of implicature set are annotated as not explicit.
Finally, the inter-annotator agreement (i.e. the proportion of queries with $3$ consistent answers and with those that have less than $3$ answers filtered) for Q1 is $79.10\%$ and Q2 is $62.11\%$, which shows consensus in Q1 while some conflicts in Q2 probably due to its subjectivity.

\input{floats/tab_human}

\input{floats/tab_eval}

\subsection{Evaluation Results}\label{sec:eval_res}

We report results on $3$ intent encoders. (1) \textbf{paraphrase} is a vanilla Sentence-BERT~\cite{reimers-gurevych-2019-sentence} model\footnote{\url{https://huggingface.co/sentence-transformers/paraphrase-mpnet-base-v2}} which shows strong performance on intent detection (used as an initialization in \citet{sung2023pretraining}). (2) \textbf{IAE}~\cite{sung2023pretraining} first employs an intent-role labeler to extract pseudo intent names and then optimizes contrastive objective leveraging utterances, pseudo intent names and golden intent names. (3) \textbf{Instructor}~\cite{su-etal-2023-one} pre-trains on multi-task dataset with instructions (``\textit{Represent the purpose for retrieval: }'') pretended on each input text. We experiment with both the base and large versions.

The evaluation results in Table~\ref{tab:eval} shows
(1) success rates from $T_{hard}$ are consistently low ($<25\%$) indicating positive utterances (i.e. utterances with the same intent) are further away that negations. In contrast, success rates for $T_{easy}$ are much higher, highlighting implicature and negation utterances are far away from one another;
(2) for binary classification, performances on the implicature split are lower than those on the original set showcasing the former is set of utterances are further awar for the expected intent than the latter;
(3) for clustering and classification, performances on implicature sets are consistently lower than the original sets, however, this might be due to the multi-labeled nature of the implicature utterances. In Appendix~\ref{sec:app_class_upper}, we verify the upper bound with \texttt{gpt-4-0613} on 10-shot classification task and demonstrates there is still a large room for improvement;
% (4) IAE improves over paraphrase on binary classification, clustering and classification. In contrast, its performance on triplet task is slightly degraded;
(4) \texttt{instructor-large} consistently improves upon its base model, and \texttt{instructor-base} that possesses a similar parameter size with IAE seems to outperform the latter on most tasks. Given this, we use \texttt{instructor-large} as the baseline embedding model for improvement.

%% file: floats/tab_human.tex
\begin{table}[t]
    \centering
    \scalebox{0.8}{
    \begin{tabular}{c|c|c}
        \specialrule{0.1em}{0.2em}{0.2em}
        \multicolumn{3}{c}{Q1: convey intent?} \\
        \hline
        original & negation & implicature \\
        \hline
        59 / 59 (100\%) & 9 / 59 (15.25\%) & 52 / 59 (88.14\%) \\
        \specialrule{0.1em}{0.2em}{0.2em}
    \end{tabular}
    }
    \\
    \scalebox{0.8}{
    \begin{tabular}{c|c|c}
        \specialrule{0.1em}{0.2em}{0.2em}
        \multicolumn{3}{c}{Q2: explicitly conveyed?} \\
        \hline
        original & negation & implicature \\
        \hline
        46 / 52 (88.46\%) & - & 15 / 40 (37.50\%) \\
        \specialrule{0.1em}{0.2em}{0.2em}
    \end{tabular}
    }
    \caption{\small Human evaluation of the quality of the automatically generated negation and implicature utterances. We note that negation mostly do not convey the original intent while implicature do (as expected). In addition, from Q2, we note that the implicature utterances are challenging (mostly implicit).}
    \label{tab:eval_human_breakdown}
    \vspace{-2mm}
\end{table}

%% file: floats/tab_eval.tex
\begin{table*}[t]
\centering
\scalebox{0.67}{
\begin{tabular}{lccccccccccccccc}
\toprule
 & \multicolumn{4}{c}{\bf Original} & \multicolumn{11}{c}{\bf Intent Semantic Toolkit} \\[0.3em]
Model & \multicolumn{2}{c}{Clustering} & \multicolumn{2}{c}{Multi-class} & \multicolumn{2}{c}{Triplet (Ori-Ori)} & \multicolumn{2}{c}{Triplet (Ori-Imp)} & \multicolumn{3}{c}{Binary Classification} & \multicolumn{2}{c}{Clustering} & \multicolumn{2}{c}{Multi-class} \\
 & KM & Agg & 0-shot & 10-shot & $T_{hard}$ & $T_{easy}$ & $T_{hard}$ & $T_{easy}$ & Ori & Imp & Neg & KM & Agg & 0-shot & 10-shot \\
\cmidrule(lr){2-5}\cmidrule(lr){6-16}
paraphrase & 81.7 & 83.5 & 61.1 & 83.3 & 22.6 & 84.8 & \bf 3.9 & 68.9 & 77.6 & 57.3 & 82.4 & 57.2 & 58.8 & 22.2 & 28.2 \\
IAE & 83.4 & 84.7 & 66.6 & 84.7 & \bf 24.0 & 84.3 & 3.2 & 67.3 & 86.6 & 70.1 & 79.6 & 58.3 & 59.9 & 25.4 & 30.1 \\
instructor-base & 83.8 & 84.9 & 67.5 & 85.8 & 19.1 & 86.1 & 2.0 & 68.0 & 89.1 & 67.3 & 78.4 & 57.9 & 59.2 & 26.2 & 30.9 \\
instructor-large & \bf 84.3 & \bf 86.0 & \bf 67.6 & \bf 86.2 & 23.4 & \bf 87.5 & 3.6 & \bf 71.0 & \bf 89.6 & \bf 73.5 & \bf 87.4 & \bf 59.1 & \bf 61.4 & \bf 28.8 & \bf 34.3 \\
\bottomrule
\end{tabular}
}
% \caption{Evaluation results with $4$ popular intent embedding models on our Intent Semantics Toolkit.}
\caption{\small 
Model performance on the original tasks and our proposed Intent Semantic Toolkit with $4$ popular intent encoders. Despite promising performance on the originaldatasets, our toolkit reveals a lack of understanding on negation and implicature.}
\label{tab:eval}
\vspace{-3mm}
\end{table*}

%% file: sections/5_train.tex
\section{Model Improvement}\label{sec:train}

We seek to improve the semantic understanding of embeddings models to negation and implicature. To answer this, we first introduce a data curation procedure where new utterances are generated based on an unlabeled dialogue corpus (\S\ref{sec:train_data}) and then investigate continued fine-tuning approaches with a contrastive learning loss objective adapted for the triplet task (\S\ref{sec:train_obj}). The lower part of \autoref{fig:framework} given a diagrammatic overview.

\input{floats/tab_main_large_test}
\input{floats/tab_main_large_dev_small}

\subsection{Fine-tune Data Curation}\label{sec:train_data}

Similar to \cite{sung2023pretraining}, we first collect a set $252,744$ unique and unlabelled utterances from general domain dialogue datasets: MultiWOZ~\cite{zang-etal-2020-multiwoz}, SGD~\cite{rastogi2020towards}, TOP~\cite{gupta-etal-2018-semantic-parsing} and TOPv2~\cite{chen-etal-2020-low}. We then use a performant LLM, namely \texttt{falcon-40b-instruct}~\cite{falcon40b} for the sub-modules in Algorithm~\ref{alg:intent_extract}.\footnote{We opt out of using OpenAI models due to their restrictions on distilling data for model training, see \S\ref{sec:lim} for details.} Another rationale for these design choices is that they ensure a separation between training and evaluation data, thereby demonstrating the model's generative capabilities.

As highlighted in Algorithm~\ref{alg:intent_extract}, for each unlabelled utterance, we first use the LLM to generate a user goal using the prompt {\em what does a customer want by saying `$u_i$'?} Then, we use a dependency parser\footnote{\url{https://spacy.io/}} to extract the action ({\tt ROOT}) or object ({\tt dobj}) tokens given the output goal phrase. If the object is not found, we prompt the LLM to summarize an object token. The final tuple is then used as a prefix for the utterance generation (see Figure~\ref{fig:extract_intent} for an example) that is eventually used for finetuning.

% To make the original uttearnces more ameable for data generation, we first extract an action-object pair for each unlabeled utterance and use the action as an intent to provide the LLM an intermediate reasoning cue that facilitates the final prompt design for generation. As show in The procedure is an alternation between LLM prompting and a dependency parser. We start with generating user goals by asking the LLM ``what does a customer want by saying [UTTERANCE]''. Next, a dependency parser\footnote{\url{https://spacy.io/}} is utilized to find action (``ROOT'') or object tokens (``dobj''). If the object is not found, we further prompt the LLM to summarize the object token. The outputs from intent extraction will be prepended in the utterance-generation prompts. See Algorithm~\ref{alg:intent_extract} for illustration. Additionally see Figure~\ref{fig:extract_intent} in Appendix for detailed prompts and examples.
% See Figure~\ref{fig:train_data_intent} for more details.

\input{floats/alg_extract_intent}

To further generate hard positive/negative utterances, we propose to utilize a ``zoo'' of prompts written by experts that asks the LLM to imagine itself as the customer and generate hard positives or negatives. And then at each time, we sample two prompts, one for generating positive and one for generating negative. \autoref{tab:train_data_prompt} in Appendix gives an overview of all the prompts we used for fine-tuning in this work. The generated utterances will be diversified before training by switching common phrases such as ``want to''. Apart from that, we also ``retrieve'' one positive utterance that has the same action-object pairs and one negative utterance that has different action-object pairs but close in cosine distance\footnote{We always use the same embedding model for utterance retrieving and fine-tuning.} See more details in Appendix~\ref{sec:app_finetune_details}.

\subsection{Fine-tuning Objective}\label{sec:train_obj}

We adopt fine-tuning objective from \citet{su-etal-2023-one,zhang2023clusterllm} where our input is a batch $\mathcal{B}$ of triplets 
$\{u,u^p,u^n\}$ generated in the previous section (\S\ref{sec:train_data}) and placed within the same instruction template used in the triplet task evaluation (in \S\ref{sec:eval_res}). More precisely, if we let $i$ and $j$ denote indices of $\mathcal{B}$, the objective function pulls together the original utterance $u_i$ and it corresponding positive example $u_i^p$ while pushing away all in-batch negative $u_j^n \forall j \in \mathcal{B}$. Precisely, we consider the following loss function:%
% $\{t_i=\{u_i,u_i^p,u_i^n\}\}_{i\in\mathcal{B}}$, where $\mathcal{B}$ is a batch of indices where each utterance is prepended with the same instruction in \S\ref{sec:eval_res}. Before training, we will first pre-process the dataset to acquire $u_i^p$ and $u_i^n$ for each $u_i$, by concatenating LLM-generated utterances and retrieved utterances and then uniformly sample one positive and one negaitve from them. The objective attempts to pull together $u_i$ and $u_i^p$, while pushing away $u_i$ and all in-batch negatives $\{u_j^n\}_{j\in\mathcal{B}}$.
\begin{equation}
\small
    \begin{split}
        l_i&=\frac{\exp(s(f(u_i),f(u_i^p))/\gamma)}{\sum_{j\in\mathcal{B}} \exp(s(f(u_i),f(u_j^n))/\gamma)} \\&+ \frac{\exp(s(f(u_i^p),f(u_i))/\gamma)}{\sum_{j\in\mathcal{B}} \exp(s(f(u_i^p),f(u_j^n))/\gamma)}
        \label{eq:train_obj}
    \end{split}
\end{equation}%
where $\gamma$ is a temperature parameter. The second term swaps the positive and anchor.

%% file: floats/tab_main_large_test.tex
\begin{table*}[t]
\centering
\scalebox{0.72}{
\begin{tabular}{lccccccccccccccc}
% \toprule
%  & \multicolumn{7}{c}{\bf Original} & \multicolumn{8}{c}{\bf Intent Semantic Toolkit} \\[0.3em]
% Model & \multicolumn{2}{c}{Triplet} & Binary & \multicolumn{2}{c}{Clustering} & \multicolumn{2}{c}{Multi-class} & \multicolumn{2}{c}{Triplet} & \multicolumn{2}{c}{Binary} & \multicolumn{2}{c}{Clustering} & \multicolumn{2}{c}{Multi-class} \\
%  & $T_{hard}$ & $T_{easy}$ & Ori & KM & Agg & 0-shot & 10-shot & $T_{hard}$ & $T_{easy}$ & Imp & Neg & KM & Agg & 0-shot & 10-shot \\[0.3em]
% \cmidrule(lr){2-8}\cmidrule(lr){9-16}
% Baseline & 23.4 & 87.5 & 89.6 & 84.3 & 86.0 & 67.6 & 86.2 & 3.3 & 70.6 & \bf 73.9 & 79.4 & 62.2 & 64.4 & 29.2 & 34.1 \\
% Disable LLM & 39.3 & 86.2 & 89.2 & 84.1 & 85.0 & 71.7 & 85.5 & 8.0 & 50.8 & 53.0 & \bf 84.8 & 65.5 & 66.9 & 31.7 & 34.3 \\
% Ours best & \bf 51.1 & \bf 93.7 & \bf 94.0 & \bf 84.6 & \bf 86.8 & \bf 73.4 & \bf 87.2 & \bf 20.4 & \bf 77.6 & 73.6 & 83.1 & \bf 65.9 & \bf 68.2 & \bf 33.9 & \bf 37.2 \\
% \bottomrule
% \\
\toprule
 & \multicolumn{4}{c}{\bf Original} & \multicolumn{11}{c}{\bf Intent Semantic Toolkit} \\[0.3em]
Model & \multicolumn{2}{c}{Clustering} & \multicolumn{2}{c}{Multi-class} & \multicolumn{2}{c}{Triplet (Ori-Ori)} & \multicolumn{2}{c}{Triplet (Ori-Imp)} & \multicolumn{3}{c}{Binary Classification} & \multicolumn{2}{c}{Clustering} & \multicolumn{2}{c}{Multi-class} \\
 & KM & Agg & 0-shot & 10-shot & $T_{hard}$ & $T_{easy}$ & $T_{hard}$ & $T_{easy}$ & Ori & Imp & Neg & KM & Agg & 0-shot & 10-shot \\
\cmidrule(lr){2-5}\cmidrule(lr){6-16}
Baseline & 84.3 & 86.0 & 67.6 & 86.2 & 23.4 & 87.5 & 3.3 & 70.6 & 89.6 & \bf 73.9 & 79.4 & 62.2 & 64.4 & 29.2 & 34.1 \\
Disable LLM & 84.1 & 85.0 & 71.7 & 85.5 & 39.3 & 86.2 & 8.0 & 50.8 & 89.2 & 53.0 & \bf 84.8 & 65.5 & 66.9 & 31.7 & 34.3 \\
Ours best & \bf 84.6 & \bf 86.8 & \bf 73.4 & \bf 87.2 & \bf 51.1 & \bf 93.7 & \bf 20.4 & \bf 77.6 & \bf 94.0 & 73.6 & 83.1 & \bf 65.9 & \bf 68.2 & \bf 33.9 & \bf 37.2 \\
\bottomrule
\end{tabular}
}
% \caption{Main results for \texttt{instructor-large} on the test set. ``Ours best'' corresponds to ``$-P^4,N^{1,3}$'' model in Table~\ref{tab:exp_main_large_dev}. Best results for each task are bolded.}
\caption{Main results for \texttt{instructor-large} on the test set. ``Ours best'' corresponds to ``$-P^4,N^{1,3}$'' model in Table~\ref{tab:exp_main_large_dev}. Our model achieves better performance on both original and our proposed toolkit.}
\label{tab:exp_main_large_test}
\vspace{-2mm}
\end{table*}

%% file: floats/tab_main_large_dev_small.tex
\begin{table*}[t]
\centering
\scalebox{0.65}{
\begin{tabular}{lcccccccccccccccc}
\toprule
 & & \multicolumn{4}{c}{\bf Original} & \multicolumn{11}{c}{\bf Intent Semantic Toolkit} \\[0.3em]
Model & Rank & \multicolumn{2}{c}{Clustering} & \multicolumn{2}{c}{Multi-class} & \multicolumn{2}{c}{Triplet (Ori-Ori)} & \multicolumn{2}{c}{Triplet (Ori-Imp)} & \multicolumn{3}{c}{Binary Classification} & \multicolumn{2}{c}{Clustering} & \multicolumn{2}{c}{Multi-class} \\
 & & KM & Agg & 0-shot & 10-shot & $T_{hard}$ & $T_{easy}$ & $T_{hard}$ & $T_{easy}$ & Ori & Imp & Neg & KM & Agg & 0-shot & 10-shot \\
\cmidrule(lr){3-6}\cmidrule(lr){7-17}
Baseline & 14 & 84.0 & 85.4 & 67.3 & 86.1 & 22.5 & 87.9 & 4.1 & 72.1 & 90.1 & 73.2 & 78.9 & 61.6 & 63.1 & 25.9 & 28.3 \\
All~$\{P^*, N^*\}$ & 7 & \cellcolor[HTML]{9AFF99}84.6 & \cellcolor[HTML]{9AFF99}85.8 & \cellcolor[HTML]{9AFF99}69.9 & \cellcolor[HTML]{9AFF99}86.5 & \cellcolor[HTML]{9AFF99}36.9 & \cellcolor[HTML]{FFCCC9}83.7 & \cellcolor[HTML]{9AFF99}12.6 & \cellcolor[HTML]{FFCCC9}56.0 & \cellcolor[HTML]{FFCCC9}73.9 & \cellcolor[HTML]{FFCCC9}35.8 & \cellcolor[HTML]{9AFF99}90.7 & \cellcolor[HTML]{9AFF99}64.2 & \cellcolor[HTML]{9AFF99}66.1 & \cellcolor[HTML]{9AFF99}27.7 & \cellcolor[HTML]{9AFF99}33.2 \\
$-P^{4},N^{1,3}$ & \textbf{1} & \cellcolor[HTML]{9AFF99}84.7 & \cellcolor[HTML]{9AFF99}86.7 & \cellcolor[HTML]{9AFF99}73.5 & \cellcolor[HTML]{9AFF99}87.8 & \cellcolor[HTML]{9AFF99}53.3 & \cellcolor[HTML]{9AFF99}94.9 & \cellcolor[HTML]{9AFF99}20.8 & \cellcolor[HTML]{9AFF99}78.4 & \cellcolor[HTML]{9AFF99}93.9 & \cellcolor[HTML]{9AFF99}73.9 & \cellcolor[HTML]{9AFF99}82.3 & \cellcolor[HTML]{9AFF99}66.2 & \cellcolor[HTML]{9AFF99}68.3 & \cellcolor[HTML]{9AFF99}34.2 & \cellcolor[HTML]{9AFF99}37.6 \\
$-N^*$ & 6 & \cellcolor[HTML]{9AFF99}85.8 & \cellcolor[HTML]{9AFF99}87.2 & \cellcolor[HTML]{9AFF99}70.3 & \cellcolor[HTML]{9AFF99}87.5 & \cellcolor[HTML]{FFCCC9}16.7 & \cellcolor[HTML]{FFCCC9}72.2 & \cellcolor[HTML]{FFCCC9}3.1 & \cellcolor[HTML]{FFCCC9}40.9 & \cellcolor[HTML]{FFCCC9}58.9 & \cellcolor[HTML]{FFCCC9}27.7 & \cellcolor[HTML]{9AFF99}89.7 & \cellcolor[HTML]{9AFF99}63.6 & \cellcolor[HTML]{9AFF99}65.7 & \cellcolor[HTML]{9AFF99}28.1 & \cellcolor[HTML]{9AFF99}33.4 \\
$-P^*$ & 12 & \cellcolor[HTML]{FFCCC9}80.6 & \cellcolor[HTML]{FFCCC9}81.0 & \cellcolor[HTML]{FFCCC9}66.4 & \cellcolor[HTML]{FFCCC9}83.0 & \cellcolor[HTML]{9AFF99}56.7 & \cellcolor[HTML]{9AFF99}91.2 & \cellcolor[HTML]{9AFF99}22.8 & \cellcolor[HTML]{FFCCC9}64.8 & \cellcolor[HTML]{9AFF99}90.7 & \cellcolor[HTML]{FFCCC9}62.0 & \cellcolor[HTML]{9AFF99}85.1 & \cellcolor[HTML]{9AFF99}62.6 & \cellcolor[HTML]{9AFF99}64.7 & \cellcolor[HTML]{FFCCC9}25.8 & \cellcolor[HTML]{9AFF99}30.3 \\
$-LLM$ & 15 & \cellcolor[HTML]{FFCCC9}83.6 & \cellcolor[HTML]{FFCCC9}84.4 & \cellcolor[HTML]{9AFF99}69.2 & \cellcolor[HTML]{FFCCC9}85.5 & \cellcolor[HTML]{9AFF99}41.8 & \cellcolor[HTML]{FFCCC9}87.2 & \cellcolor[HTML]{9AFF99}8.9 & \cellcolor[HTML]{FFCCC9}51.6 & \cellcolor[HTML]{FFCCC9}89.1 & \cellcolor[HTML]{FFCCC9}53.5 & \cellcolor[HTML]{9AFF99}84.0 & \cellcolor[HTML]{9AFF99}63.4 & \cellcolor[HTML]{9AFF99}65.4 & \cellcolor[HTML]{9AFF99}27.3 & \cellcolor[HTML]{9AFF99}32.1 \\
\bottomrule
\end{tabular}
}
\caption{Ablation study for various prompts with \texttt{instructor-large} on dev set. Negative sign represents disabling. For instance, ``$-P^*$'' means disabling all hard positive prompts and only using retrieved positive utterances. ``$-LLM$'' means disabling data generated from LLM and only using retrieved utterances. ``$-P^4-N^{1,3}$'' means disabling prompts $P^4$, $N^1$ and $N^3$. Numbers in the brackets of first column are the rankings~\cite{colombo2022best}. \begin{tabular}{c} \cellcolor[HTML]{9AFF99}green\end{tabular} represents increased score compared with ``Baseline'' (vanilla \texttt{instructor-large}) and \begin{tabular}{c} \cellcolor[HTML]{FFCCC9}red\end{tabular} vice versa.}
\label{tab:exp_main_large_dev_small}
\vspace{-4mm}
\end{table*}

%% file: floats/alg_extract_intent.tex
\begin{algorithm}[t]
    \small
    \KwIn{A set of unlabeled utterances $\{u_i\}_{i=1}^N$.}
    \BlankLine
    \caption{\small Intent Extraction Pipeline}\label{alg:intent_extract}
    $s \leftarrow \{\}$ \\
    \For{ i=1:N }{
        $g_i \leftarrow GoalGeneration(u_i)$ \\
        $(a_i,o_i) \leftarrow DependencyParser(g_i)$ \\
        \If{ $o_i$ is None }{
            $o_i=SummarizeObject(u_i,g_i,a_i)$
        }
        $s \leftarrow s \cup (u_i,g_i,a_i,o_i)$
    }
    \KwOut{s}
    % \vspace{-3mm}
\end{algorithm}

%% file: sections/6_exp.tex
\input{floats/tab_comparison_iae}
% switch order
% retrieval baseline

\section{Experimental Results}\label{sec:exp}
In this section, we first evaluate our fine-tuned models in \S\ref{sec:train} on our proposed Intent Semantics Toolkit. We then select the best prompt combination for training models on the validation set, and then evaluate on the test set. Finally, we shed lights on the correlations between tasks in \S\ref{sec:exp_corr} and the effectiveness of LLM-augmented utterances when combining intent-aware encoder training~\cite{sung2023pretraining} in \S\ref{sec:exp_aug}.

\subsection{Experimental Setup}\label{sec:exp_setup}
We follow fine-tuning parameters in \citet{su-etal-2023-one}, except that we choose smaller learning rate $4\times 10^{-6}$, batch size $8$, maximum sequence length $128$ and train with one epoch. These parameters are consistent across all experiments in this paper.

\subsection{Main Results}\label{sec:exp_main}

The fine-tuned models are evaluated on our proposed Intent Semantics Toolkit in Section~\ref{sec:eval}. We split the evaluation data into $50\%$ for dev and $50\%$ for test to select proper prompt combinations. We report results based on the test set using \texttt{instructor-large} as the baseline in \autoref{tab:exp_main_large_test} and highlight some ablations study with various prompts in \autoref{tab:exp_main_large_dev_small} (see Appendix for experiments with other prompts and \texttt{instructor-base}). As It is difficult to exhaust all possible combinations of prompts, we consider a subset of experiments in our ablations and rank the various trained models using \citet{colombo2022best}.
We observe that (1) disabling all LLM-generated data achieves lower performance than {\em Baseline},
(2) disabling LLM-generated hard positive degrades performances on clustering and classification tasks, while disabling hard negative degrades performances on triplet tasks and binary classification tasks, and (3) the best performance is achieved by ``$-P^4-N^{1,3}$'', which outperforms the {\em Baseline} on most tasks. As seen in \autoref{tab:exp_main_large_test}, while we notice consistent improvement with our model across original and our proposed tasks, the magnitude of improvements is not uniform (eg. a $\uparrow 27.7$ on  $T_{hard}$ {\em vs.} $\downarrow 0.3$ on the binary task with implicatures).  To understand this better, we consider the correlation between tasks.

\subsection{Correlation between tasks}\label{sec:exp_corr}
\begin{figure}[t]
    \centering
    \includegraphics[width=0.48\textwidth]{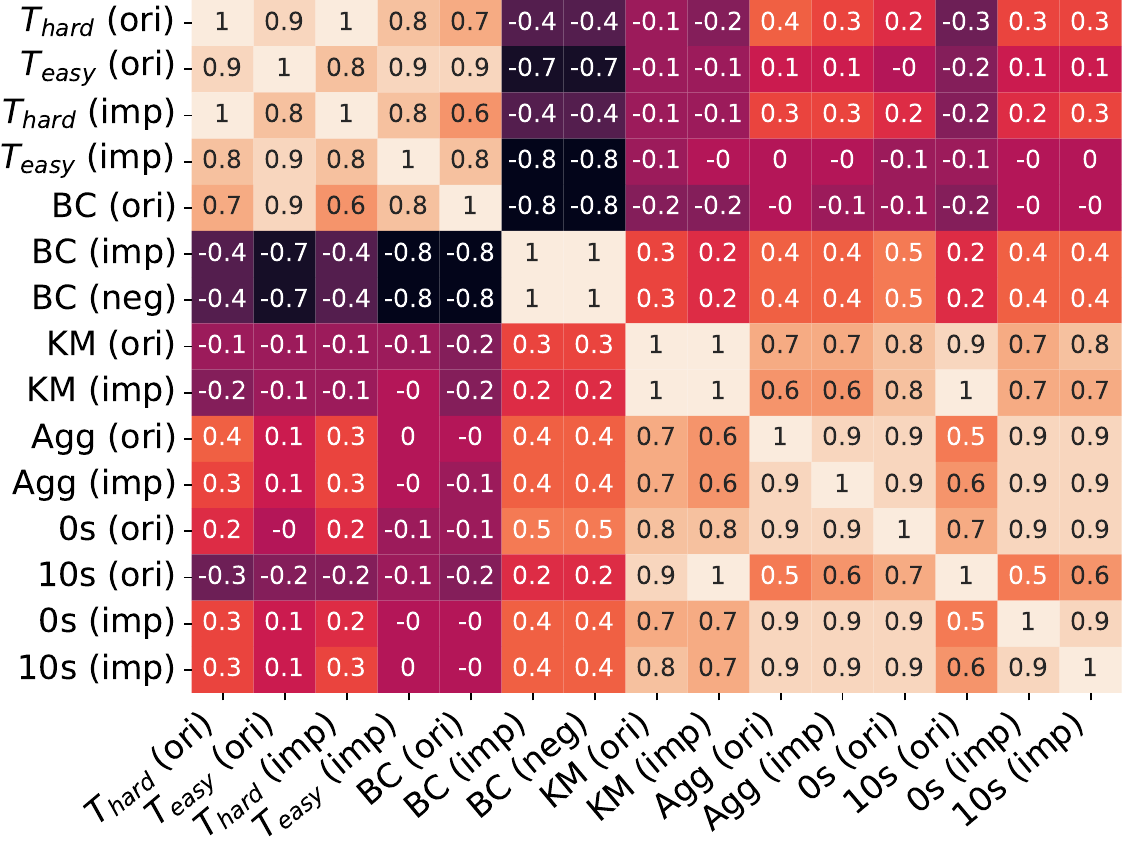}
    \caption{Pearson correlations between tasks using performances on dev set as a feature vector for each task.}
    \label{fig:exp_corr}
    \vspace{-6mm}
\end{figure}
We plot the pearson correlation matrix between pairs of tasks across \texttt{instructor-large} (\autoref{tab:exp_main_large_dev_small}) and \texttt{instructor-base} (\autoref{tab:exp_main_base_dev}) dev set performances. \autoref{fig:exp_corr} clearly highlights the negative correlation between the first two tasks (triplet tasks and binary classification) and the last two (clustering and classification). Thus, it may be important to consider trade-offs in improving an embedding model across all tasks that we leave as future work.

\subsection{Augment Intent-Aware-Encoder}\label{sec:exp_aug}

We further show the performances by adding LLM-generated data into Intent-Aware-Encoder (IAE) pre-training data in order to properly compare performances.
% We use contrastive objective for IAE and IAE-pos, and switch to Eq.~\ref{eq:train_obj} for IAE-pos-neg since it simultaneously accomodates hard positive and negative examples. We observe significant performance improvements on triplet tasks and binary classification when using ``IAE-pos-neg'' and comparable performances for the other two tasks.
``IAE-pos'' replaces part of the IAE loss function that uses pseudo label names with LLM generated data augmentation (for positive labels), while ``IAE-pos-neg'' uses both positive and negative examples (either retrieved or LLM-generated) in the contrastive loss term.
% We show further discussions in Appendix~\ref{sec:}
% Thus it confirms the capability of LLM-generated data (especially hard negatives) on improving semantic understanding even when the gold labels are accessible.

%% file: floats/tab_comparison_iae.tex
\begin{table*}[ht!]
\centering
\scalebox{0.65}{
\begin{tabular}{lccccccccccccccc}
\toprule
 & \multicolumn{4}{c}{\bf Original} & \multicolumn{11}{c}{\bf Intent Semantic Toolkit} \\[0.3em]
Model & \multicolumn{2}{c}{Clustering} & \multicolumn{2}{c}{Multi-class} & \multicolumn{2}{c}{Triplet (Ori-Ori)} & \multicolumn{2}{c}{Triplet (Ori-Imp)} & \multicolumn{3}{c}{Binary Classification} & \multicolumn{2}{c}{Clustering} & \multicolumn{2}{c}{Multi-class} \\
 & KM & Agg & 0-shot & 10-shot & $T_{hard}$ & $T_{easy}$ & $T_{hard}$ & $T_{easy}$ & Ori & Imp & Neg & KM & Agg & 0-shot & 10-shot \\
\cmidrule(lr){2-5}\cmidrule(lr){6-16}
IAE                    & 83.4             & 84.6            & 66.3             & 84.5             & 25.3           & 84.9           & 3.5            & 68.1           & 87.3        & 70.5        & 80.0       & 61.9             & 64.2            & 25.5             & 30.4             \\
IAE-pos                & \cellcolor[HTML]{9AFF99}83.6             & \cellcolor[HTML]{9AFF99}85.0            & \cellcolor[HTML]{FFCCC9}66.1             & \cellcolor[HTML]{FFCCC9}84.4             & \cellcolor[HTML]{9AFF99}27.7           & \cellcolor[HTML]{9AFF99}85.8           & \cellcolor[HTML]{9AFF99}4.3            & \cellcolor[HTML]{9AFF99}68.9           & \cellcolor[HTML]{9AFF99}87.8        & \cellcolor[HTML]{9AFF99}70.9        & \cellcolor[HTML]{9AFF99}80.9       & \cellcolor[HTML]{9AFF99}62.1             & \cellcolor[HTML]{9AFF99}64.5            & \cellcolor[HTML]{9AFF99}26.2             & \cellcolor[HTML]{9AFF99}30.9             \\
IAE-pos-neg            & \cellcolor[HTML]{FFCCC9}83.0             & \cellcolor[HTML]{FFCCC9}84.2            & \cellcolor[HTML]{FFCCC9}65.6             & \cellcolor[HTML]{FFCCC9}84.4             & \cellcolor[HTML]{9AFF99}40.6           & \cellcolor[HTML]{9AFF99}90.8           & \cellcolor[HTML]{9AFF99}13.7           & \cellcolor[HTML]{9AFF99}78.4           & \cellcolor[HTML]{9AFF99}91.3        & \cellcolor[HTML]{9AFF99}76.1        & \cellcolor[HTML]{9AFF99}81.5       & \cellcolor[HTML]{9AFF99}62.1             & \cellcolor[HTML]{9AFF99}65.1            & \cellcolor[HTML]{9AFF99}25.6             & \cellcolor[HTML]{9AFF99}30.6  \\
\bottomrule
\end{tabular}
}
\caption{Adding LLM generated hard positive/negative to Intent-Aware-Encoder (IAE)~\cite{sung2023pretraining} pre-training. Results averaged over $3$ random seeds.}
\label{tab:compare_iae}
\vspace{-4mm}
\end{table*}

%% file: sections/7_con.tex
\section{Conclusion and Future Works}\label{sec:con}

In this paper, we propose a new evaluation toolkit for intent embedding models that measure their semantic understanding on two lingusitic phenomenon common in conversational systems-- negation and implicature. For this we propose a novel triplet task, a binary classification task and challenge test splits that evaluate the model's semantic understanding on downstream intent recognition (classification and clustering) tasks.  Our study shows that current intent embedding models do not have sufficient understanding of these two real world phenomenon, i.e. negation and implicature. We then propose to integrate hard positives and negatives generated from an LLM with a ``zoo'' of prompts to fine-tune the model. The fine-tuning is conducted on a set of unlabeled utterances from general domain and is evaluated on our proposed toolkit. Our best model demonstrates improvements across most tasks over baseline models. We further combine LLM-generated data with Intent-Aware-Encoder training and show performance improvements across all original test datasets and most newly introduced evaluation datasets in the Intent Semantic Toolkit. Finally, the correlation analysis between pairs of tasks indicates that a better balance between negation-related tasks and regular benchmark tasks need to be achieved. Our work also inspires future works to develop and evaluate an instruction-following embedding model that can improve performance via prompting without the need for further fine-tuning of the model.

%% file: sections/lim.tex
\section*{Limitations}\label{sec:lim}

In this paper, we used ChatGPT for evaluation and a smaller LLM \texttt{falcon-40b-instruct} for training. This is mainly due to the legal concern. Our use of OpenAI for this publication, is to the best our knowledge, in compliance with applicable OpenAI’s terms and conditions as of March 14, 2023. Our results show that Falcon-based data augmentation improves performance significantly. However, we argue that using a larger and more capable LLM can potentially improve the quality of data generated and thus further improve the model.

\section*{Ethical Considerations}\label{sec:ethics}

This paper uses various open-source datasets and models for evaluation and training, which are reproducible. Our evaluation toolkit uses OpenAI models which might impact the reproducibility. However, we set the temperature to $0$ in order to reduce the variances.

%% file: sections/ack.tex
\section*{Acknowledgements}
We would like to thank James Gung for insightful discussions. We also want to thank Hossein Aboutalebi and Dennis Ulmer for their efforts in contributing to our human evaluation study.

%% file: sections/app.tex
\begin{figure*}
    \centering
    \includegraphics[width=0.99\textwidth]{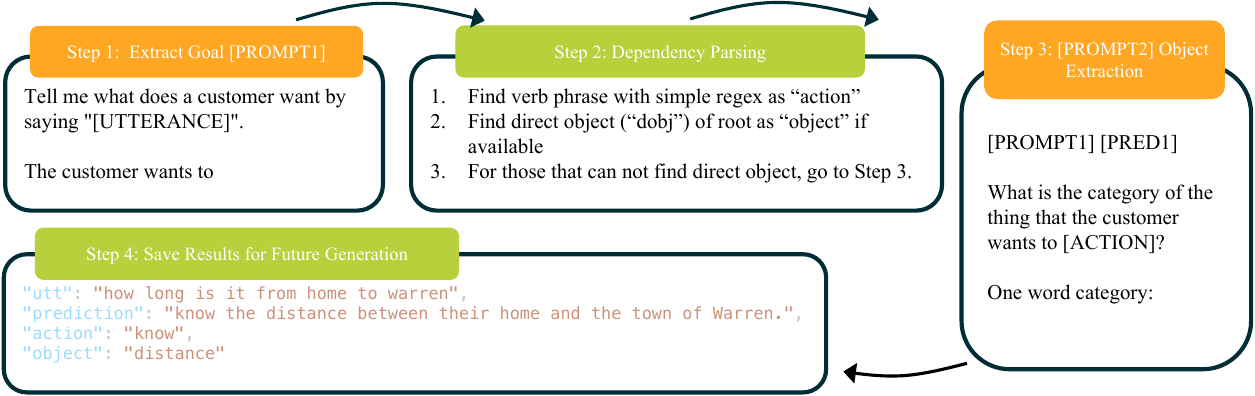}
    \caption{Intent extraction pipeline. We first prompt the LLM for generating the user goal from the utterance. We then find the action-object pair from the generated goal. For those that can not find objects during second step, we will further summarize the object with LLM. And finally, we will save the goal and action-object pair for further generation.}
    \label{fig:extract_intent}
\end{figure*}

% \section{Human Annotation Guidelines}\label{sec:app_human_guide}

\begin{figure*}
    \centering
    \begin{subfigure}
        \centering
        \includegraphics[width=0.48\linewidth]{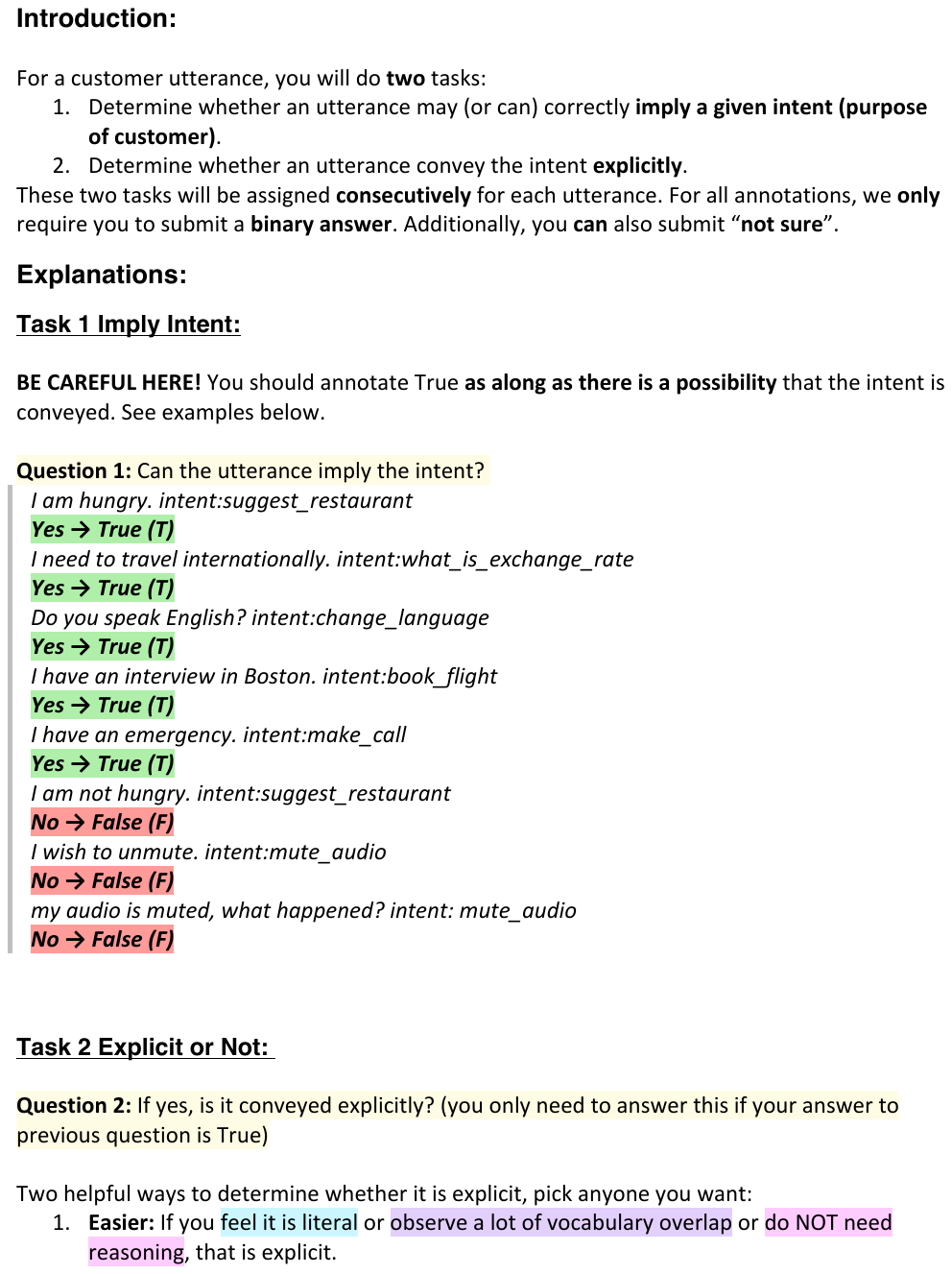}
    \end{subfigure}
    \begin{subfigure}
        \centering
        \includegraphics[width=0.48\linewidth]{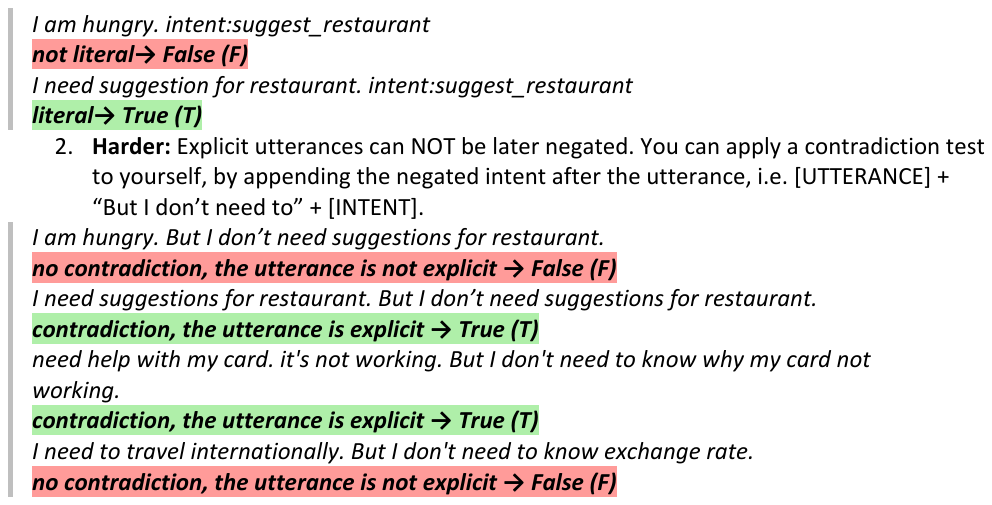}
    \end{subfigure}
    \caption{Human annotation guidelines for quality control. Our annotators are from diverse cultures and ethical groups including Asians, Europeans and Americans.}
    \label{tab:app_annotation}
\end{figure*}

\section{Classification Upper-Bound with GPT-4}\label{sec:app_class_upper}
In Section~\ref{sec:eval}, we discussed the multi-label nature of implicature utterances because of the inherent ambiguity of these utterances.
For instance, ``\textit{set the mood please}'' could mean both ``INTENT:change\_light\_hue'' and ``INTENT:play\_music''. Such a phenomena could be one of the reasons that the evaluated embedding models show degraded performances compared with original set. In order to decompose the effects of multi-label, we present a classification upper-bound on 10-shot multi-class classification  with \texttt{gpt-4-0613}. We present \texttt{gpt-4-0613} with the utterance and the top-$5$ predictions from \texttt{instructor-large} plus one ground truth label. We then ask the model to classify it into one of the class. The calculated accuracy is comparable with $10$-shot classification performance in Table~\ref{tab:eval}. The obtained results from \texttt{gpt-4-0613} is $55.2\%$ which is significantly higher than $34.3\%$. This further illustrates that we still have a lot of rooms for improvement for embedding models.

\section{More Details on Fine-tuning: Retrieving and Diversifying}\label{sec:app_finetune_details}

\noindent \textbf{Retrieving} Apart from LLM-generated data, we also use the retrieved utterances which is considered as a standard data collecting procedure (and thus a baseline method) in contrastive learning. We retrieve one positive and one negative data for each of the utterances in the unlabeled corpus. For positive one, we simply uniformly choose one utterance from the same-intent utterance group. Notice that although there is no guarantee a positive utterance can be retrieved, in practice we found most of utterances in the training corpus can find positive pairs, \textit{i.e.} $95.39\%$ ($241,080$ utterances). And we simply filter out the rest of utterances that can not retrieve at least one positive. For negative one, we first encode all the negative utterances and the original utterance into the embedding space with the same embedding model that is going to be trained. And then we calculate cosine distance between original embedding and negative embeddings, and sort them from smaller to larger. We then acquire the one in the middle of the list as we empirically found that these utterances are both similar to the original utterances in terms of surface forms and possess different intents.

\noindent \textbf{Diversifying} We empirically observe that the generated utterances usually contain similar surface forms, \textit{e.g.} most utterances from $P^4$ starts with ``i do not want to''. This is due to the question forms being used in the specific prompts, and may potentially harm the diversity of training data. In order to diversify them, we manually create more patterns such ``i try not to'', ``i prefer not to'' for $P^4$, and then modify the original utterance with it. In addition, we also remove those responses that reject to produce answers by identifying keywords ``ai language model''.

\noindent \textbf{Computing Budget} All our models can be trained on a single 48GB GPU with less than 3 hours of training.

% \section{Discussion on Augmenting Intent-Aware-Encoder}

\section{Qualitative Analysis on Errors}
Note that we have some qualitative examples in Figure~\ref{fig:challenges} that are sampled from our dataset to demonstrate the characteristics of implicature and negation. In order to better showcase some examples, we provide more qualitative analysis here on CLINC150, especially failure cases during binary classification tasks (classify to one of original label or negated label) using \texttt{instructor-large}.

Implicature errors:

\begin{quote}
\textit{i've recently retired and i'm trying to cut back on my spending} (credit\_limit\_change) \\
\textit{the passenger will be stopping in london, is the time difference going to be a problem?} (timezone) \\
\textit{i think my bags missed the connection} (lost\_luggage) \\
\textit{i'm having a hard time understanding the app because it's not in my native language} (change\_language) \\
\textit{i can't decide what to have for dinner} (flip\_coin)
\end{quote}

Negation errors:

\begin{quote}
\textit{i didn't mean to ask for the spanish word for pasta} (no\_need\_to\_request\_translate)\\
\textit{please don't change my name, it was a mistake} (no\_need\_to\_change\_user\_name)\\
\textit{my visa was damaged but it has been replaced} (card\_not\_damaged)\\
\textit{i like this song, don't skip it} (stay\_at\_current\_song)\\
\textit{i mistakenly set an alarm} (no\_need\_to\_set\_alarm)
\end{quote}

Furthermore, we performed clustering on these failure cases with \texttt{instructor-large} embeddings and then perform $k$-means clustering. We then run Tf-idf with each cluster as a document to acquire feature vectors for them, where the high value entries indicate keywords for the cluster. We show some example keywords for both negation and implicature here:

Implicature:
'my', 'to', 'the', 'you', 'your', 'can', 'not'
'my', 'card', 'credit', 'to', 'the', 'wallet', 'bank'
'to', 'the', 'for', 'recipe', 'salad', 'lunch', 'dinner'
'to', 'the', 'my', 'need', 'meeting', 'have', 'for'
'my', 'to', 'meter', 'the', 'utility', 'reading', 'bill'
'the', 'song', 'to', 'my', 'this', 'quiet', 'is'
'to', 'the', 'car', 'my', 'need', 'get', 'long'
'to', 'in', 'thinking', 'the', 'for', 'trip', 'my'
'language', 'in', 'my', 'the', 'to', 'app', 'english'
'my', 'luggage', 'the', 'suitcase', 'bags', 'bag', 'plane'

Negation:
'reservations', 'list', 'to', 'for', 'the'
'phone', 'to', 'know', 'my', 'you'
'account', 'my', 'bank', '401k', 'paid'
'happened', 'song', 'what', 'the', 'is'
'whisper', 'mode', 'settings', 'changed', 'unexpectedly'
'card', 'credit', 'my', 'the', 'limit'
'meeting', 'booked', 'my', 'flight', 'to'
'thanks', 'for', 'not', 'nothing', 'thank'
'my', 'engine', 'light', 'to', 'check'
'jump', 'car', 'start', 'my', 'to'

The results did not indicate strong bias towards a specific topic, and thus showing that such failures are ubiquitous across domains and intent classes.

\input{floats/tab_main_large_dev}

\input{floats/tab_main_base_test}

\input{floats/tab_main_base_dev}

\input{floats/tab_hphnprompt}

%% file: floats/tab_main_large_dev.tex
\begin{table*}[t]
\centering
\scalebox{0.65}{
\begin{tabular}{lccccccccccccccccc}
\toprule
 & & \multicolumn{4}{c}{\bf Original} & \multicolumn{11}{c}{\bf Intent Semantic Toolkit} \\[0.3em]
Model & Rank & \multicolumn{2}{c}{Clustering} & \multicolumn{2}{c}{Multi-class} & \multicolumn{2}{c}{Triplet (Ori-Ori)} & \multicolumn{2}{c}{Triplet (Ori-Imp)} & \multicolumn{3}{c}{Binary Classification} & \multicolumn{2}{c}{Clustering} & \multicolumn{2}{c}{Multi-class} \\
 & & KM & Agg & 0-shot & 10-shot & $T_{hard}$ & $T_{easy}$ & $T_{hard}$ & $T_{easy} $& Ori & Imp & Neg & KM & Agg & 0-shot & 10-shot \\
\cmidrule(lr){3-6}\cmidrule(lr){7-17}
Baseline & 14 & 84.0 & 85.4 & 67.3 & 86.1 & 22.5 & 87.9 & 4.1 & 72.1 & 90.1 & 73.2 & 78.9 & 61.6 & 63.1 & 25.9 & 28.3 \\
All~$\{P^*, N^*\}$ & 7 & 84.6 & 85.8 & 69.9 & 86.5 & \cellcolor[HTML]{9AFF99}36.9 & \cellcolor[HTML]{FFCCC9}83.7 & \cellcolor[HTML]{9AFF99}12.6 & \cellcolor[HTML]{FFCCC9}56.0 & \cellcolor[HTML]{FFCCC9}73.9 & \cellcolor[HTML]{FFCCC9}35.8 & \cellcolor[HTML]{9AFF99}90.7 & 64.2 & 66.1 & 27.7 & 33.2 \\
~$-P^{4},N^{1,3}$ & \textbf{1} & \cellcolor[HTML]{9AFF99}84.7 & \cellcolor[HTML]{9AFF99}86.7 & \cellcolor[HTML]{9AFF99}73.5 & \cellcolor[HTML]{9AFF99}87.8 & \cellcolor[HTML]{9AFF99}53.3 & \cellcolor[HTML]{9AFF99}94.9 & \cellcolor[HTML]{9AFF99}20.8 & \cellcolor[HTML]{9AFF99}78.4 & \cellcolor[HTML]{9AFF99}93.9 & \cellcolor[HTML]{9AFF99}73.9 & \cellcolor[HTML]{9AFF99}82.3 & \cellcolor[HTML]{9AFF99}66.2 & \cellcolor[HTML]{9AFF99}68.3 & \cellcolor[HTML]{9AFF99}34.2 & \cellcolor[HTML]{9AFF99}37.6 \\
~$-P^{1,4},N^{1,3}$ & 2 & \cellcolor[HTML]{9AFF99}85.2 & \cellcolor[HTML]{9AFF99}86.0 & \cellcolor[HTML]{9AFF99}73.6 & \cellcolor[HTML]{9AFF99}87.2 & \cellcolor[HTML]{9AFF99}54.4 & \cellcolor[HTML]{9AFF99}94.5 & \cellcolor[HTML]{9AFF99}20.5 & \cellcolor[HTML]{9AFF99}76.4 & \cellcolor[HTML]{9AFF99}93.3 & \cellcolor[HTML]{FFCCC9}71.4 & \cellcolor[HTML]{9AFF99}83.3 & \cellcolor[HTML]{9AFF99}66.1 & \cellcolor[HTML]{9AFF99}68.0 & \cellcolor[HTML]{9AFF99}33.8 & \cellcolor[HTML]{9AFF99}37.5 \\
~$-P^{4},N^{1}$ & 3 & \cellcolor[HTML]{9AFF99}84.8 & \cellcolor[HTML]{9AFF99}85.5 & \cellcolor[HTML]{9AFF99}72.7 & \cellcolor[HTML]{9AFF99}87.0 & \cellcolor[HTML]{9AFF99}55.2 & \cellcolor[HTML]{9AFF99}94.7 & \cellcolor[HTML]{9AFF99}21.1 & \cellcolor[HTML]{9AFF99}79.3 & \cellcolor[HTML]{9AFF99}94.2 & \cellcolor[HTML]{9AFF99}77.2 & \cellcolor[HTML]{9AFF99}82.5 & \cellcolor[HTML]{9AFF99}66.5 & \cellcolor[HTML]{9AFF99}67.8 & \cellcolor[HTML]{9AFF99}33.9 & \cellcolor[HTML]{9AFF99}36.9 \\
~$-P^{1,2,4},N^{1,3}$ & 4 & \cellcolor[HTML]{9AFF99}84.9 & \cellcolor[HTML]{9AFF99}85.4 & \cellcolor[HTML]{9AFF99}73.0 & \cellcolor[HTML]{9AFF99}86.9 & \cellcolor[HTML]{9AFF99}56.6 & \cellcolor[HTML]{9AFF99}94.2 & \cellcolor[HTML]{9AFF99}21.8 & \cellcolor[HTML]{9AFF99}75.3 & \cellcolor[HTML]{9AFF99}93.2 & \cellcolor[HTML]{FFCCC9}71.5 & \cellcolor[HTML]{9AFF99}83.5 & \cellcolor[HTML]{9AFF99}65.9 & \cellcolor[HTML]{9AFF99}67.7 & \cellcolor[HTML]{9AFF99}33.0 & \cellcolor[HTML]{9AFF99}36.8 \\
~$-P^{4},N^{3}$ & 5 & \cellcolor[HTML]{9AFF99}85.2 & \cellcolor[HTML]{9AFF99}86.0 & \cellcolor[HTML]{9AFF99}73.2 & \cellcolor[HTML]{9AFF99}86.8 & \cellcolor[HTML]{9AFF99}52.4 & \cellcolor[HTML]{9AFF99}94.3 & \cellcolor[HTML]{9AFF99}18.8 & \cellcolor[HTML]{9AFF99}75.8 & \cellcolor[HTML]{9AFF99}92.9 & \cellcolor[HTML]{FFCCC9}71.1 & \cellcolor[HTML]{9AFF99}83.6 & \cellcolor[HTML]{9AFF99}65.7 & \cellcolor[HTML]{9AFF99}67.7 & \cellcolor[HTML]{9AFF99}33.1 & \cellcolor[HTML]{9AFF99}36.9 \\
~$-N^*$ & 6 & \cellcolor[HTML]{9AFF99}85.8 & \cellcolor[HTML]{9AFF99}87.2 & \cellcolor[HTML]{9AFF99}70.3 & \cellcolor[HTML]{9AFF99}87.5 & \cellcolor[HTML]{FFCCC9}16.7 & \cellcolor[HTML]{FFCCC9}72.2 & \cellcolor[HTML]{FFCCC9}3.1 & \cellcolor[HTML]{FFCCC9}40.9 & \cellcolor[HTML]{FFCCC9}58.9 & \cellcolor[HTML]{FFCCC9}27.7 & \cellcolor[HTML]{9AFF99}89.7 & \cellcolor[HTML]{9AFF99}63.6 & \cellcolor[HTML]{9AFF99}65.7 & \cellcolor[HTML]{9AFF99}28.1 & \cellcolor[HTML]{9AFF99}33.4 \\
~$-P^4$ & 8 & \cellcolor[HTML]{9AFF99}84.2 & \cellcolor[HTML]{FFCCC9}84.8 & \cellcolor[HTML]{9AFF99}72.2 & \cellcolor[HTML]{9AFF99}86.2 & \cellcolor[HTML]{9AFF99}54.8 & \cellcolor[HTML]{9AFF99}94.0 & \cellcolor[HTML]{9AFF99}20.7 & \cellcolor[HTML]{9AFF99}77.7 & \cellcolor[HTML]{9AFF99}93.1 & \cellcolor[HTML]{FFCCC9}72.2 & \cellcolor[HTML]{9AFF99}83.3 & \cellcolor[HTML]{9AFF99}65.9 & \cellcolor[HTML]{9AFF99}67.2 & \cellcolor[HTML]{9AFF99}33.7 & \cellcolor[HTML]{9AFF99}36.7 \\
~$-P^2$ & 9 & \cellcolor[HTML]{9AFF99}84.5 & \cellcolor[HTML]{FFCCC9}85.1 & \cellcolor[HTML]{9AFF99}72.7 & \cellcolor[HTML]{9AFF99}86.4 & \cellcolor[HTML]{9AFF99}37.4 & \cellcolor[HTML]{FFCCC9}82.8 & \cellcolor[HTML]{9AFF99}13.1 & \cellcolor[HTML]{FFCCC9}53.4 & \cellcolor[HTML]{FFCCC9}70.8 & \cellcolor[HTML]{FFCCC9}32.7 & \cellcolor[HTML]{9AFF99}90.9 & \cellcolor[HTML]{9AFF99}65.8 & \cellcolor[HTML]{9AFF99}67.8 & \cellcolor[HTML]{9AFF99}34.0 & \cellcolor[HTML]{9AFF99}36.6 \\
~$-P^1$ & 10 & \cellcolor[HTML]{9AFF99}84.9 & \cellcolor[HTML]{FFCCC9}85.3 & \cellcolor[HTML]{9AFF99}73.4 & \cellcolor[HTML]{9AFF99}87.1 & \cellcolor[HTML]{9AFF99}36.0 & \cellcolor[HTML]{FFCCC9}82.9 & \cellcolor[HTML]{9AFF99}12.0 & \cellcolor[HTML]{FFCCC9}53.0 & \cellcolor[HTML]{FFCCC9}68.3 & \cellcolor[HTML]{FFCCC9}31.5 & \cellcolor[HTML]{9AFF99}92.0 & \cellcolor[HTML]{9AFF99}65.7 & \cellcolor[HTML]{9AFF99}67.6 & \cellcolor[HTML]{9AFF99}33.9 & \cellcolor[HTML]{9AFF99}36.7 \\
~$-P^4,N^2$ & 11 & \cellcolor[HTML]{9AFF99}84.3 & \cellcolor[HTML]{FFCCC9}85.2 & \cellcolor[HTML]{9AFF99}72.6 & \cellcolor[HTML]{9AFF99}86.6 & \cellcolor[HTML]{9AFF99}32.4 & \cellcolor[HTML]{FFCCC9}86.0 & \cellcolor[HTML]{9AFF99}6.3 & \cellcolor[HTML]{FFCCC9}55.0 & \cellcolor[HTML]{FFCCC9}86.6 & \cellcolor[HTML]{FFCCC9}51.3 & \cellcolor[HTML]{9AFF99}86.6 & \cellcolor[HTML]{9AFF99}65.7 & \cellcolor[HTML]{9AFF99}67.5 & \cellcolor[HTML]{9AFF99}33.4 & \cellcolor[HTML]{9AFF99}36.2 \\
~$-P^*$ & 12 & \cellcolor[HTML]{FFCCC9}80.6 & \cellcolor[HTML]{FFCCC9}81.0 & \cellcolor[HTML]{FFCCC9}66.4 & \cellcolor[HTML]{FFCCC9}83.0 & \cellcolor[HTML]{9AFF99}56.7 & \cellcolor[HTML]{9AFF99}91.2 & \cellcolor[HTML]{9AFF99}22.8 & \cellcolor[HTML]{FFCCC9}64.8 & \cellcolor[HTML]{9AFF99}90.7 & \cellcolor[HTML]{FFCCC9}62.0 & \cellcolor[HTML]{9AFF99}85.1 & \cellcolor[HTML]{9AFF99}62.6 & \cellcolor[HTML]{9AFF99}64.7 & \cellcolor[HTML]{FFCCC9}25.8 & \cellcolor[HTML]{9AFF99}30.3 \\
~$-P^3$ & 13 & \cellcolor[HTML]{FFCCC9}83.4 & \cellcolor[HTML]{FFCCC9}84.5 & \cellcolor[HTML]{9AFF99}72.5 & \cellcolor[HTML]{9AFF99}86.3 & \cellcolor[HTML]{9AFF99}39.0 & \cellcolor[HTML]{FFCCC9}82.7 & \cellcolor[HTML]{9AFF99}13.5 & \cellcolor[HTML]{FFCCC9}53.2 & \cellcolor[HTML]{FFCCC9}68.9 & \cellcolor[HTML]{FFCCC9}32.9 & \cellcolor[HTML]{9AFF99}91.3 & \cellcolor[HTML]{9AFF99}65.2 & \cellcolor[HTML]{9AFF99}66.6 & \cellcolor[HTML]{9AFF99}33.3 & \cellcolor[HTML]{9AFF99}35.4 \\
~$-LLM$ & 15 & \cellcolor[HTML]{FFCCC9}83.6 & \cellcolor[HTML]{FFCCC9}84.4 & \cellcolor[HTML]{9AFF99}69.2 & \cellcolor[HTML]{FFCCC9}85.5 & \cellcolor[HTML]{9AFF99}41.8 & \cellcolor[HTML]{FFCCC9}87.2 & \cellcolor[HTML]{9AFF99}8.9 & \cellcolor[HTML]{FFCCC9}51.6 & \cellcolor[HTML]{FFCCC9}89.1 & \cellcolor[HTML]{FFCCC9}53.5 & \cellcolor[HTML]{9AFF99}84.0 & \cellcolor[HTML]{9AFF99}63.4 & \cellcolor[HTML]{9AFF99}65.4 & \cellcolor[HTML]{9AFF99}27.3 & \cellcolor[HTML]{9AFF99}32.1 \\
\bottomrule
\end{tabular}
}
\caption{This is the complete version of Table~\ref{tab:exp_main_large_dev_small} that includes all the prompt variants.}
% Ablation study for various prompts with \texttt{instructor-large} on dev set. Negative sign represents disabling. For instance, ``$-P^*$'' means disabling all hard positive prompts and only using retrieved positive utterances. ``$-LLM$'' means disabling data generated from LLM and only using retrieved utterances. ``$-P^4-N^{1,3}$'' means disabling prompts $P^4$, $N^1$ and $N^3$. Numbers in the brackets of first column are the rankings~\cite{colombo2022best}. \begin{tabular}{c} \cellcolor[HTML]{9AFF99}green\end{tabular} represents increased score and \begin{tabular}{c} \cellcolor[HTML]{FFCCC9}red\end{tabular} vice versa compared with ``Baseline'', i.e. vanilla \texttt{instructor-large}.}
\label{tab:exp_main_large_dev}
\end{table*}

%% file: floats/tab_main_base_test.tex
\begin{table*}[t]
\centering
\scalebox{0.66}{
\begin{tabular}{lccccccccccccccc}
% \toprule
%  & \multicolumn{7}{c}{\bf Original} & \multicolumn{8}{c}{\bf Intent Semantic Toolkit} \\[0.3em]
% Model & \multicolumn{2}{c}{Triplet} & Binary & \multicolumn{2}{c}{Clustering} & \multicolumn{2}{c}{Multi-class} & \multicolumn{2}{c}{Triplet} & \multicolumn{2}{c}{Binary} & \multicolumn{2}{c}{Clustering} & \multicolumn{2}{c}{Multi-class} \\
%  & $T_{hard}$ & $T_{easy}$ & ori & KM & Agg & 0-shot & 10-shot & $T_{hard}$ & $T_{easy}$ & imp & neg & KM & Agg & 0-shot & 10-shot \\
% \cmidrule(lr){2-8}\cmidrule(lr){9-16}
% Baseline               & 19.1           & 86.1           & 89.1        & \bf 83.8             & 84.9            & 67.5             & 85.8             & 2.0            & 67.6           & 67.0        & 78.4       & 61.1             & 62.9            & 26.4             & 31.1             \\
% Disable LLM            & 24.5           & 80.5           & 83.6        & 82.6             & 84.2            & \bf 69.5             & 85.3             & 3.0            & 43.4           & 43.1        & \bf 83.6       & 63.1             & 65.1            & 27.0             & 31.7             \\
% Ours best              & \bf 46.9           & \bf 93.3           & \bf 94.1        & 83.3             & \bf 85.2            & 69.3             & \bf 86.4             & \bf 17.8           & \bf 80.9           & \bf 81.0        & 78.9       & \bf 63.5             & \bf 65.5            & \bf 28.3             & \bf 33.8 \\ 
% \bottomrule
% \\
\toprule
 & \multicolumn{4}{c}{\bf Original} & \multicolumn{11}{c}{\bf Intent Semantic Toolkit} \\[0.3em]
Model & \multicolumn{2}{c}{Clustering} & \multicolumn{2}{c}{Multi-class} & \multicolumn{2}{c}{Triplet (Ori-Ori)} & \multicolumn{2}{c}{Triplet (Ori-Imp)} & \multicolumn{3}{c}{Binary Classification} & \multicolumn{2}{c}{Clustering} & \multicolumn{2}{c}{Multi-class} \\
 & KM & Agg & 0-shot & 10-shot & $T_{hard}$ & $T_{easy}$ & $T_{hard}$ & $T_{easy}$ & Ori & Imp & Neg & KM & Agg & 0-shot & 10-shot \\
\cmidrule(lr){2-5}\cmidrule(lr){6-16}
Baseline & \bf 83.8 & 84.9 & 67.5 & 85.8 & 19.1 & 86.1 & 2.0 & 67.6 & 89.1 & 67.0 & 78.4 & 61.1 & 62.9 & 26.4 & 31.1 \\
Disable LLM & 82.6 & 84.2 & \bf 69.5 & 85.3 & 24.5 & 80.5 & 3.0 & 43.4 & 83.6 & 43.1 & \bf 83.6 & 63.1 & 65.1 & 27.0 & 31.7 \\
Ours best & 83.3 & \bf 85.2 & 69.3 & \bf 86.4 & \bf 46.9 & \bf 93.3 & \bf 17.8 & \bf 80.9 & \bf 94.1 & \bf 81.0 & 78.9 & \bf 63.5 & \bf 65.5 & \bf 28.3 & \bf 33.8 \\
\bottomrule
\end{tabular}
}
\caption{Main results for \texttt{instructor-base} on test set. ``Ours best'' corresponds to ``-$P^4,N^{1,3}$'' model in Table~\ref{tab:exp_main_base_dev}. Best results for each task are bolded.}
\label{tab:exp_main_base_test}
\end{table*}

%% file: floats/tab_main_base_dev.tex
\begin{table*}[t]
\centering
\scalebox{0.64}{
\begin{tabular}{lcccccccccccccccc}
\toprule
 & & \multicolumn{4}{c}{\bf Original} & \multicolumn{11}{c}{\bf Intent Semantic Toolkit} \\[0.3em]
Model & Rank & \multicolumn{2}{c}{Clustering} & \multicolumn{2}{c}{Multi-class} & \multicolumn{2}{c}{Triplet (Ori-Ori)} & \multicolumn{2}{c}{Triplet (Ori-Imp)} & \multicolumn{3}{c}{Binary Classification} & \multicolumn{2}{c}{Clustering} & \multicolumn{2}{c}{Multi-class} \\
 & & KM & Agg & 0-shot & 10-shot & $T_{hard}$ & $T_{easy}$ & $T_{hard}$ & $T_{easy} $& Ori & Imp & Neg & KM & Agg & 0-shot & 10-shot \\
\cmidrule(lr){3-6}\cmidrule(lr){7-17}
Baseline      & 14         & 83.5             & 84.5            & 67.3             & 86.1             & 18.9           & 86.5           & 2.6            & 68.2           & 89.0        & 67.6        & 78.0       & 61.6             & 63.1            & 25.9             & 28.3             \\
All $\{P^*$, $N^*\}$     & 5         & \cellcolor[HTML]{9AFF99} 83.8             & \cellcolor[HTML]{FFCCC9} 84.4            & \cellcolor[HTML]{9AFF99} 69.9             & \cellcolor[HTML]{9AFF99} 86.5             & \cellcolor[HTML]{9AFF99} 31.1           & \cellcolor[HTML]{FFCCC9} 82.1           & \cellcolor[HTML]{9AFF99} 9.4            & \cellcolor[HTML]{FFCCC9} 55.9           & \cellcolor[HTML]{FFCCC9} 67.0        & \cellcolor[HTML]{FFCCC9} 32.8        & \cellcolor[HTML]{9AFF99} 87.6       & \cellcolor[HTML]{9AFF99} 64.2             & \cellcolor[HTML]{9AFF99} 66.1            & \cellcolor[HTML]{9AFF99} 27.7             & \cellcolor[HTML]{9AFF99} 33.2             \\
~$-P^4,N^{1,3}$  & \bf 1           & \cellcolor[HTML]{FFCCC9} 83.5             & \cellcolor[HTML]{9AFF99} 84.8            & \cellcolor[HTML]{9AFF99} 69.8             & \cellcolor[HTML]{9AFF99} 86.5             & \cellcolor[HTML]{9AFF99} 48.6           & \cellcolor[HTML]{9AFF99} 94.0           & \cellcolor[HTML]{9AFF99} 18.2           & \cellcolor[HTML]{9AFF99} 81.4           & \cellcolor[HTML]{9AFF99} 94.1        & \cellcolor[HTML]{9AFF99} 81.8        & \cellcolor[HTML]{9AFF99} 78.2       & \cellcolor[HTML]{9AFF99} 64.0             & \cellcolor[HTML]{9AFF99} 66.1            & \cellcolor[HTML]{9AFF99} 28.2             & \cellcolor[HTML]{9AFF99} 33.9             \\
~$-P^4,N^1$  & 2             & \cellcolor[HTML]{FFCCC9} 82.8             & \cellcolor[HTML]{FFCCC9} 83.9            & \cellcolor[HTML]{9AFF99} 69.2             & \cellcolor[HTML]{9AFF99} 86.3             & \cellcolor[HTML]{9AFF99} 49.5           & \cellcolor[HTML]{9AFF99} 93.5           & \cellcolor[HTML]{9AFF99} 19.5           & \cellcolor[HTML]{9AFF99} 79.3           & \cellcolor[HTML]{9AFF99} 92.9        & \cellcolor[HTML]{9AFF99} 77.5        & \cellcolor[HTML]{9AFF99} 79.1       & \cellcolor[HTML]{9AFF99} 64.5             & \cellcolor[HTML]{9AFF99} 66.3            & \cellcolor[HTML]{9AFF99} 28.0             & \cellcolor[HTML]{9AFF99} 33.5             \\
~$-P^{1,4},N^{1,3}$  & 3         & \cellcolor[HTML]{FFCCC9} 83.1             & \cellcolor[HTML]{FFCCC9} 84.3            & \cellcolor[HTML]{9AFF99} 70.0             & \cellcolor[HTML]{9AFF99} 86.4             & \cellcolor[HTML]{9AFF99} 48.1           & \cellcolor[HTML]{9AFF99} 93.6           & \cellcolor[HTML]{9AFF99} 18.0           & \cellcolor[HTML]{9AFF99} 79.0           & \cellcolor[HTML]{9AFF99} 93.6        & \cellcolor[HTML]{9AFF99} 77.2        & \cellcolor[HTML]{9AFF99} 78.7       & \cellcolor[HTML]{9AFF99} 64.1             & \cellcolor[HTML]{9AFF99} 65.9            & \cellcolor[HTML]{9AFF99} 27.5             & \cellcolor[HTML]{9AFF99} 33.2             \\
~$-P^4,N^3$  & 4             & \cellcolor[HTML]{9AFF99} 83.8             & \cellcolor[HTML]{FFCCC9} 84.2            & \cellcolor[HTML]{9AFF99} 69.4             & \cellcolor[HTML]{FFCCC9} 85.9             & \cellcolor[HTML]{9AFF99} 48.1           & \cellcolor[HTML]{9AFF99} 93.9           & \cellcolor[HTML]{9AFF99} 17.5           & \cellcolor[HTML]{9AFF99} 80.2           & \cellcolor[HTML]{9AFF99} 93.4        & \cellcolor[HTML]{9AFF99} 79.4        & \cellcolor[HTML]{9AFF99} 78.7       & \cellcolor[HTML]{9AFF99} 63.5             & \cellcolor[HTML]{9AFF99} 65.9            & \cellcolor[HTML]{9AFF99} 27.6             & \cellcolor[HTML]{9AFF99} 33.2             \\
~$-P^{1,2,4},N^{1,3}$   & 6      & \cellcolor[HTML]{FFCCC9} 83.2             & \cellcolor[HTML]{FFCCC9} 84.2            & \cellcolor[HTML]{9AFF99} 69.7             & \cellcolor[HTML]{9AFF99} 86.1             & \cellcolor[HTML]{9AFF99} 47.8           & \cellcolor[HTML]{9AFF99} 92.7           & \cellcolor[HTML]{9AFF99} 16.5           & \cellcolor[HTML]{9AFF99} 75.9           & \cellcolor[HTML]{9AFF99} 92.7        &  \cellcolor[HTML]{9AFF99} 73.0      &  \cellcolor[HTML]{9AFF99} 79.6      & \cellcolor[HTML]{9AFF99} 63.5             & \cellcolor[HTML]{9AFF99} 66.4            & \cellcolor[HTML]{9AFF99} 27.7             & \cellcolor[HTML]{9AFF99} 33.1         \\
~$-N^*$         & 7           & \cellcolor[HTML]{9AFF99} 84.0             & \cellcolor[HTML]{9AFF99} 85.6            & \cellcolor[HTML]{9AFF99} 70.3             & \cellcolor[HTML]{9AFF99} 87.5             & \cellcolor[HTML]{FFCCC9} 10.7           & \cellcolor[HTML]{FFCCC9} 64.5           & \cellcolor[HTML]{FFCCC9} 1.2            & \cellcolor[HTML]{FFCCC9} 32.8           & \cellcolor[HTML]{FFCCC9} 49.3        & \cellcolor[HTML]{FFCCC9} 19.8        & \cellcolor[HTML]{9AFF99} 91.0       & \cellcolor[HTML]{9AFF99} 63.6             & \cellcolor[HTML]{9AFF99} 65.7            & \cellcolor[HTML]{9AFF99} 28.1             & \cellcolor[HTML]{9AFF99} 33.4             \\
~$-P^4$      & 8             & \cellcolor[HTML]{FFCCC9} 81.9             & \cellcolor[HTML]{FFCCC9} 83.3            & \cellcolor[HTML]{9AFF99} 69.0             & \cellcolor[HTML]{FFCCC9} 85.9             & \cellcolor[HTML]{9AFF99} 48.2           & \cellcolor[HTML]{9AFF99} 93.2           & \cellcolor[HTML]{9AFF99} 17.8           & \cellcolor[HTML]{9AFF99} 78.6           & \cellcolor[HTML]{9AFF99} 92.3        & \cellcolor[HTML]{9AFF99} 74.7        & \cellcolor[HTML]{9AFF99} 79.6       & \cellcolor[HTML]{9AFF99} 64.0             & \cellcolor[HTML]{9AFF99} 65.8            & \cellcolor[HTML]{9AFF99} 27.4             & \cellcolor[HTML]{9AFF99} 32.8             \\
~$-P^2$       & 9            & \cellcolor[HTML]{FFCCC9} 83.1             & \cellcolor[HTML]{FFCCC9} 83.8            & \cellcolor[HTML]{9AFF99} 69.6             & \cellcolor[HTML]{FFCCC9} 85.9             & \cellcolor[HTML]{9AFF99} 32.5           & \cellcolor[HTML]{FFCCC9} 81.9           & \cellcolor[HTML]{9AFF99} 9.7            & \cellcolor[HTML]{FFCCC9} 53.7           & \cellcolor[HTML]{FFCCC9} 68.0        & \cellcolor[HTML]{FFCCC9} 32.6        & \cellcolor[HTML]{9AFF99} 87.5       & \cellcolor[HTML]{9AFF99} 64.0             & \cellcolor[HTML]{9AFF99} 66.0            & \cellcolor[HTML]{9AFF99} 27.6             & \cellcolor[HTML]{9AFF99} 32.7             \\
~$-P^1$       & 10            & \cellcolor[HTML]{FFCCC9} 82.9             & \cellcolor[HTML]{FFCCC9} 84.0            & \cellcolor[HTML]{9AFF99} 69.8             & \cellcolor[HTML]{9AFF99} 86.2             & \cellcolor[HTML]{9AFF99} 31.9           & \cellcolor[HTML]{FFCCC9} 81.2           & \cellcolor[HTML]{9AFF99} 9.4            & \cellcolor[HTML]{FFCCC9} 54.3           & \cellcolor[HTML]{FFCCC9} 64.8        & \cellcolor[HTML]{FFCCC9} 30.3        & \cellcolor[HTML]{9AFF99} 88.2       & \cellcolor[HTML]{9AFF99} 63.6             & \cellcolor[HTML]{9AFF99} 65.8            & \cellcolor[HTML]{9AFF99} 27.6             & \cellcolor[HTML]{9AFF99} 32.6             \\
~$-P^4,N^2$  & 11             & \cellcolor[HTML]{FFCCC9} 83.4             & \cellcolor[HTML]{FFCCC9} 83.7            & \cellcolor[HTML]{9AFF99} 69.8             & \cellcolor[HTML]{9AFF99} 86.1             & \cellcolor[HTML]{FFCCC9} 19.6           & \cellcolor[HTML]{FFCCC9} 78.6           & \cellcolor[HTML]{FFCCC9} 2.3            & \cellcolor[HTML]{FFCCC9} 43.9           & \cellcolor[HTML]{FFCCC9} 79.0        & \cellcolor[HTML]{FFCCC9} 38.4        & \cellcolor[HTML]{9AFF99} 85.6       & \cellcolor[HTML]{9AFF99} 63.1             & \cellcolor[HTML]{9AFF99} 66.4            & \cellcolor[HTML]{9AFF99} 27.4             & \cellcolor[HTML]{9AFF99} 32.7             \\
~$-P^3$       & 12            & \cellcolor[HTML]{FFCCC9} 83.3             & \cellcolor[HTML]{FFCCC9} 83.7            & \cellcolor[HTML]{9AFF99} 69.8             & \cellcolor[HTML]{FFCCC9} 85.9             & \cellcolor[HTML]{9AFF99} 31.4           & \cellcolor[HTML]{FFCCC9} 79.7           & \cellcolor[HTML]{9AFF99} 9.4            & \cellcolor[HTML]{FFCCC9} 51.6           & \cellcolor[HTML]{FFCCC9} 61.0        & \cellcolor[HTML]{FFCCC9} 27.0        & \cellcolor[HTML]{9AFF99} 88.7       & \cellcolor[HTML]{9AFF99} 64.0             & \cellcolor[HTML]{9AFF99} 65.9            & \cellcolor[HTML]{9AFF99} 27.4             & \cellcolor[HTML]{9AFF99} 32.6             \\
~$-P^*$           & 13         & \cellcolor[HTML]{FFCCC9} 80.0             & \cellcolor[HTML]{FFCCC9} 80.0            & \cellcolor[HTML]{FFCCC9} 66.4             & \cellcolor[HTML]{FFCCC9} 83.0             & \cellcolor[HTML]{9AFF99} 49.5           & \cellcolor[HTML]{9AFF99} 90.6           & \cellcolor[HTML]{9AFF99} 18.1           & \cellcolor[HTML]{FFCCC9} 67.6           & \cellcolor[HTML]{9AFF99} 90.0        & \cellcolor[HTML]{FFCCC9} 64.5        & \cellcolor[HTML]{9AFF99} 80.3       & \cellcolor[HTML]{9AFF99} 62.6             & \cellcolor[HTML]{9AFF99} 64.7            & \cellcolor[HTML]{FFCCC9} 25.8             & \cellcolor[HTML]{9AFF99} 30.3             \\
~$-LLM$          & 15         & \cellcolor[HTML]{FFCCC9} 82.8             & \cellcolor[HTML]{FFCCC9} 83.3            & \cellcolor[HTML]{9AFF99} 69.2             & \cellcolor[HTML]{FFCCC9} 85.5             & \cellcolor[HTML]{9AFF99} 26.0           & \cellcolor[HTML]{FFCCC9} 81.3           & \cellcolor[HTML]{9AFF99} 3.4            & \cellcolor[HTML]{FFCCC9} 44.9           & \cellcolor[HTML]{FFCCC9} 83.9        & \cellcolor[HTML]{FFCCC9} 44.7        & \cellcolor[HTML]{9AFF99} 83.2       & \cellcolor[HTML]{9AFF99} 63.4             & \cellcolor[HTML]{9AFF99} 65.4            & \cellcolor[HTML]{9AFF99} 27.3             & \cellcolor[HTML]{9AFF99} 32.1             \\
\bottomrule
\end{tabular}
}
\caption{Ablation study for various prompts with \texttt{instructor-base} on dev set. See Table~\ref{tab:exp_main_large_dev} for descriptions.}
\label{tab:exp_main_base_dev}
\end{table*}

%% file: floats/tab_hphnprompt.tex
\begin{table*}[t]
    \centering
    \scalebox{0.88}{
    \begin{tabularx}{\textwidth}{XX}
        \specialrule{0.1em}{0.2em}{0.2em}
         Prompt & Example \\
        \hline
        ~$P^1$: ... Tell me 3 other kinds of [OBJECT] that you want to [ACTION] ... & what are the bike lanes like around the orlando area (know about,transportation) $\rightarrow$ interested in knowing about the public transportation system in orlando. \\
        ~$P^2$: ... Give me 1 other ways to express what you want ... & can you still just recommend one please?? (be recommended,product) $\rightarrow$ can you please suggest a specific product or service that you think would be a good fit for me? \\
        ~$P^3$: ... Give me 1 reasons that you want to do this ... & is highways 72 icy this morning? (know,weather) $\rightarrow$ could you help me know if the roads are icy so that i can take necessary precautions while driving. \\
        ~$P^4$: ... Give me 2 things that you do not want to do in this scenario ... & remind me tomorrow to pick up charlie, pickup new glasses, grab groceries (be reminded,reminder) $\rightarrow$ i'd rather avoid forgetting to pick up charlie. \\
        ~$N^1$: ... Give me 2 other things you want to [ACTION] rather than [OBJECT] ... & remind me tomorrow to pick up charlie, pickup new glasses, grab groceries (be reminded,reminder) $\rightarrow$ i want to be reminded to call my mom and wish her a happy birthday. \\
        ~$N^2$: .. Now you no longer need to [ACTION] [OBJECT]. Give me 2 reasons for that ... & is highways 72 icy this morning? (know,weather) $\rightarrow$ i can check the weather forecast on my phone or computer. \\
        ~$N^3$: ... Now you do not want to [ACTION] [OBJECT], give me 2 other things you want to do ... & when is my wakeup alarm on thursday (know,time) $\rightarrow$ want to know if there are any important events happening on thursday. \\
        \specialrule{0.1em}{0.2em}{0.2em} \\
    \end{tabularx}
    }
    \vspace{-2mm}
    \caption{Hard positive/negative utterance generation prompts. The extracted intents are shown in the brackets (action,object) on the right. The arrows indicate before and after prompting.}
    \label{tab:train_data_prompt}
    \vspace{-3mm}
\end{table*}